\documentclass[journal,twoside,web]{ieeecolor}
\usepackage{generic}
\usepackage{cite}
\usepackage{amsmath,amssymb,amsfonts}
\usepackage{algorithmic}
\usepackage{graphicx}
\usepackage{algorithm,algorithmic}
\usepackage{hyperref}
\hypersetup{hidelinks=true}
\usepackage{textcomp}
\usepackage{multirow}
\usepackage{subcaption}
\usepackage{pifont}
\usepackage{xcolor}
\usepackage{tabularx}
\newcolumntype{Y}{>{\centering\arraybackslash}X}

\newcommand{\new}[0]{\textcolor{black}}

\def\BibTeX{{\rm B\kern-.05em{\sc i\kern-.025em b}\kern-.08em
    T\kern-.1667em\lower.7ex\hbox{E}\kern-.125emX}}

\begin{document}

\title{DAUNet: A Lightweight UNet Variant with Deformable Convolutions and Parameter-Free Attention for Medical Image Segmentation}

\author{Adnan Munir, Muhammad Shahid Jabbar, Shujaat Khan 
\thanks{Manuscript submitted on [Month XX, 2025]. This work was supported in part by the King Fahd University of Petroleum \& Minerals (KFUPM) under Early Career Research Grant no. EC241027 and the KFUPM Ibn Battuta Global Scholarship Program Grant No: ISP241-COE-872.}
\thanks{A. Munir is with Department of Electrical Engineering (ISY), Information Coding (ICG), Link\"{o}ping University, Link\"{o}ping, Sweden.}
\thanks{M.S. Jabbar and S. Khan are with the SDAIA-KFUPM Joint Research Center for Artificial Intelligence, King Fahd University of Petroleum \& Minerals, Dhahran 31261, Saudi Arabia.}
\thanks{S. Khan is also with the Department of Computer Engineering, College of Computing and Mathematics, King Fahd University of Petroleum \& Minerals, Dhahran 31261, Saudi Arabia (e-mail: shujaat.khan@kfupm.edu.sa).}}

\maketitle

\begin{abstract}
Medical image segmentation plays a pivotal role in automated diagnostic and treatment planning systems. In this work, we present DAUNet, a novel lightweight UNet variant that integrates Deformable V2 Convolutions and Parameter-Free Attention (SimAM) to improve spatial adaptability and context-aware feature fusion without increasing model complexity. DAUNet’s bottleneck employs dynamic deformable kernels to handle geometric variations, while the decoder and skip pathways are enhanced using SimAM attention modules for saliency-aware refinement. Extensive evaluations on two challenging datasets, FH-PS-AoP (fetal head and pubic symphysis ultrasound) and FUMPE (CT-based pulmonary embolism detection), demonstrate that DAUNet outperforms state-of-the-art models in Dice score, HD95, and ASD, while maintaining superior parameter efficiency. Ablation studies highlight the individual contributions of deformable convolutions and SimAM attention. DAUNet's robustness to missing context and low-contrast regions establishes its suitability for deployment in real-time and resource-constrained clinical environments.
\end{abstract}


\section{Introduction}

Medical image segmentation is a foundational task in computer-assisted diagnosis, enabling the precise localization and delineation of anatomical structures that are critical for clinical interpretation, surgical planning, and disease monitoring. Accurate and automated segmentation reduces manual effort and inter-observer variability, particularly in high-throughput clinical settings. Despite significant advances achieved through convolutional neural networks (CNNs), especially the widely adopted UNet architecture~\cite{unet}, key challenges persist, most notably in achieving robustness, handling anatomical variability, and maintaining computational efficiency.

Although effective in many scenarios, the classical UNet architecture presents several limitations. Its use of fixed-grid convolutions restricts adaptability to variable-sized features and irregular organ boundaries. \new{Recent deformable-convolution segmentation networks have also emphasized boundary-aware modeling to better align predictions with anatomical contours~\cite{ju2025boundary}.}
Additionally, UNet often struggles in low-contrast or noisy environments, common in modalities such as ultrasound \cite{khan2021switchable, huh2023tunable} and CT angiography, where anatomical boundaries are not clearly visible \cite{jiang2024segmentation, kang2025feasibility}. Moreover, UNet lacks mechanisms to capture long-range dependencies, which are crucial for modeling global context in complex medical images.

To overcome these shortcomings, recent works have explored enhancements to UNet via transformer-based modules and attention mechanisms \cite{fatnet,tan2024novel}. For instance, Masoudi et al.~\cite{fatnet} proposed FAT-Net, which augments a UNet-style backbone with transformer branches to capture long-range interactions and feature adaptation modules to suppress background noise. Similarly, Zhang et al.~\cite{trans} introduced TransAttUNet, incorporating a Self-Aware Attention (SAA) module that integrates Transformer Self-Attention (TSA) and Global Spatial Attention (GSA) to improve multi-scale feature fusion. Other methods such as DSEUNet~\cite{DSUNet} and MISSFormer~\cite{MFormer} attempt to bridge CNN and transformer paradigms. DSEUNet deepens the UNet backbone while introducing Squeeze-and-Excitation (SE) blocks~\cite{SENet} and hierarchical supervision. MISSFormer, on the other hand, employs enhanced transformer blocks and multi-scale fusion to balance local and global feature representation.

General-purpose models such as MedSAM~\cite{Medsam}, adapted from the Segment Anything Model (SAM), offer prompt-based segmentation across various modalities. Trained on over 1.5 million image–mask pairs, MedSAM shows strong performance on CT, MRI, and endoscopy images. However, limitations remain due to the underrepresentation of certain modalities (e.g., mammography) and imprecise vessel boundary segmentation when using bounding-box prompts.

While the aforementioned models demonstrate commendable segmentation performance, they often suffer from high computational complexity and slower inference, limiting their suitability for real-time or resource-constrained environments.

Hybrid models like H2Former~\cite{HFormer} and SCUNet++~\cite{scunetpp} further aim to unify the strengths of CNNs and transformers. H2Former leverages hierarchical token-wise and channel-wise attention to model both local and global dependencies. SCUNet++ integrates CNN bottlenecks and dense skip connections to improve pulmonary embolism (PE) segmentation. Although SCUNet++ achieves high Dice scores on PE datasets, it tends to produce blocky segmentation outputs on large lesions and has a substantial parameter burden. Other methods, such as CE-Net~\cite{CENet}, augment UNet with Dense Atrous Convolution (DAC) and Residual Multi-kernel Pooling (RMP) blocks to improve feature representation, but their multi-branch architectures increase memory requirements and limit scalability.

Motivated by the need for efficient, adaptable, and robust segmentation models suitable for real-world clinical deployment, we propose \textbf{DAUNet}, a lightweight and effective UNet-based architecture featuring two key innovations:
\begin{itemize}
    \item \textbf{Improved Bottleneck:} A lightweight deformable convolution-based bottleneck module~\cite{dai2017deformable,zhu2019deformable} that introduces dynamic, spatially adaptive receptive fields. This design enables the model to better capture geometric deformations and irregular anatomical boundaries.
    
    \item \textbf{Improved Decoder:} A parameter-free attention mechanism (SimAM)~\cite{yang2021simam} is integrated into the decoder and skips connections to enhance spatial feature representation and facilitate efficient feature fusion, without increasing model complexity.
\end{itemize}

To demonstrate its effectiveness, we evaluate \textbf{DAUNet} on two challenging medical image segmentation tasks: (1) fetal head and pubic symphysis segmentation from transperineal ultrasound using the FH-PS-AoP dataset~\cite{PSFH}, and (2) pulmonary embolism detection in CT angiography using the FUMPE dataset~\cite{fumpe}. Both tasks are characterized by substantial anatomical variability, low-contrast regions, and limited contextual information, factors that commonly impair the performance of conventional models. Through comprehensive experiments, DAUNet achieves superior segmentation accuracy, robustness to missing context, and significantly improved parameter efficiency compared to state-of-the-art methods. These results, supported by ablation and robustness analyses, highlight DAUNet’s practical potential for deployment in real-time and resource-constrained clinical environments.

The remainder of this paper is organized as follows: Section~\ref{sec:method} describes the proposed DAUNet architecture in detail. Section~\ref{sec:experiments} outlines the experimental setup, datasets, and evaluation metrics. In Section~\ref{sec:results}, we present and discuss quantitative and qualitative results, including ablation studies and robustness analysis. In Section~\ref{sec:discussion}, we provides a discussion on clinical impact. Finally, Section~\ref{sec:conclusion} concludes the paper and outlines directions for future work.

\section{Methodology}\label{sec:method}
The proposed framework is built upon the UNet architecture with two key modifications: Deformable V2 Convolutions~\cite{dai2017deformable,zhu2019deformable} in the bottleneck block and Parameter-Free Attention~\cite{yang2021simam} in the decoder block.

\subsection{Deformable Convolution}
Deformable convolution~\cite{dai2017deformable,zhu2019deformable} extends the conventional convolution operation by incorporating spatial offsets that adapt dynamically based on input features. This mechanism enables the network to focus on pertinent regions, effectively capturing spatial deformations, such as scaling, rotation, and complex anatomical structures.

\begin{figure}[ht!]
    \centering
    \includegraphics[width=1\linewidth]{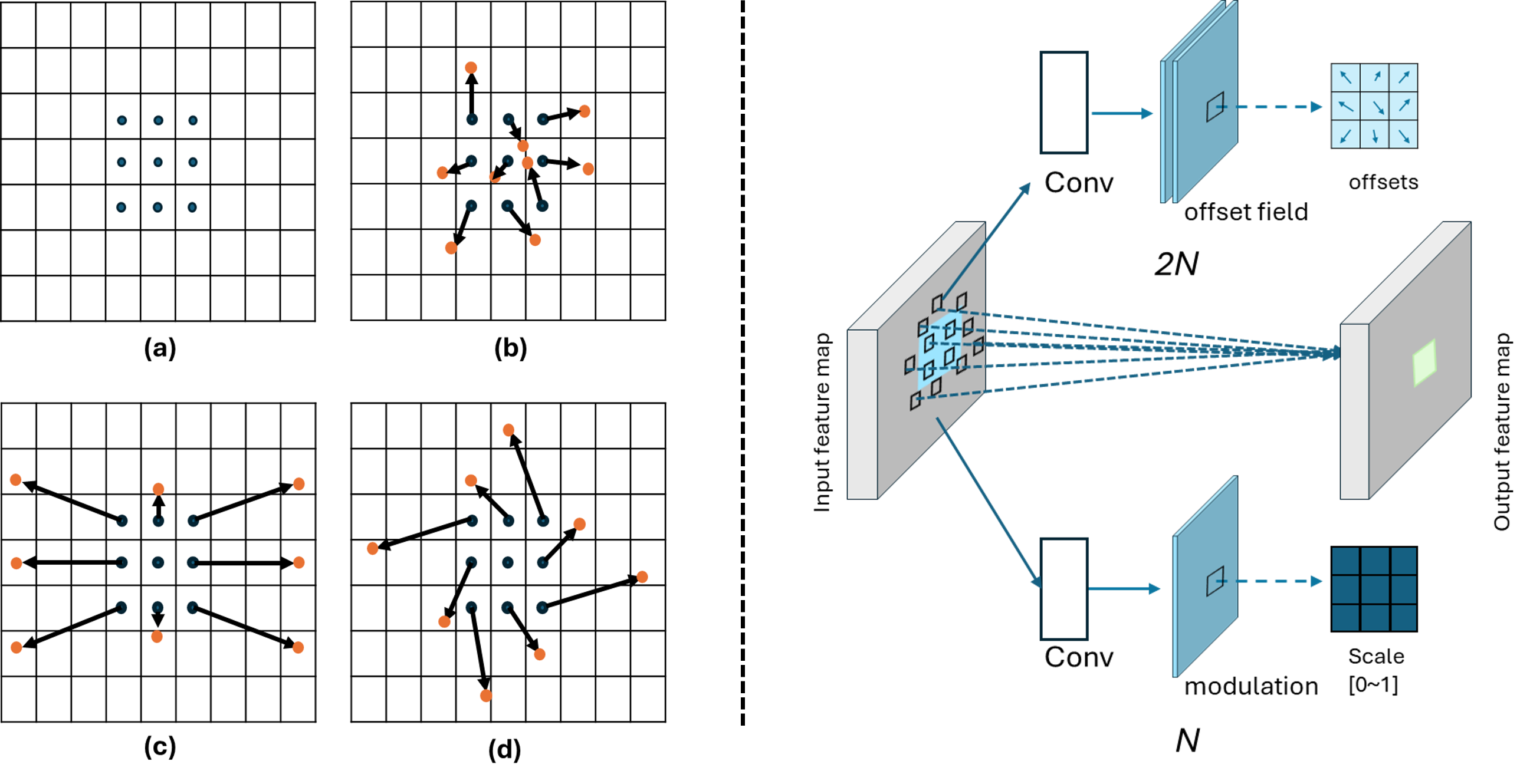}
    \caption{Comparison between standard and deformable convolutions.
    (a) Standard convolution with a fixed $3\times3$ sampling grid, where sampling locations are identical across all spatial positions.
    \new{(b) Deformable convolution with learnable offsets, where sampling locations are dynamically shifted based on input features.}
    \new{(c) Example of expanded and directionally adaptive receptive fields produced by deformable convolution, allowing the kernel to capture elongated or irregular structures.}
    \new{(d) Example of locally adaptive and anisotropic sampling patterns, demonstrating how deformable convolution aligns sampling points with underlying object geometry.}
    \new{The diagram on the right illustrates the modulated deformable convolution mechanism, where convolutional layers predict both spatial offsets and modulation scalars that jointly control sampling locations and their relative importance.}}
    \label{fig:deform_fig}
\end{figure}

\new{Figure \ref{fig:deform_fig} illustrates the operational differences between standard and deformable convolutions. The left panel depicts an illustrative comparison between $3 \times 3$ standard convolution and deformable convolution, showing how deformable convolution adapts sampling locations and receptive fields to better align with underlying geometry. In contrast, the right panel shows a deformable convolution, where each sampling location is adjusted by a learnable offset, depicted by the black arrows.}

\subsubsection{Deformable Convolution Operation}
For a given input feature map $\mathbf{F}$ of size $C \times H \times W$ and a convolution kernel $\mathbf{K}$ of size $C \times w \times w$, the output at location $\mathbf{p}$ is computed as:

\begin{equation}
Y(\mathbf{p}) = \sum_{\mathbf{k} \in \mathcal{R}} \mathbf{K}(\mathbf{k}) \cdot \mathbf{F}(\mathbf{p} + \mathbf{k} + \Delta \mathbf{p_k}),
\end{equation}

where $\mathcal{R}$ is the receptive field, and $\Delta \mathbf{p_k}$ is the learnable offset. Since $\mathbf{p} + \mathbf{k} + \Delta \mathbf{p_k}$ may be a non-integer location, bilinear interpolation is applied to approximate the feature values at fractional coordinates.

\subsubsection{Modulated Deformable Convolution}
To further enhance flexibility, a modulation scalar $\alpha_k \in [0, 1]$ is introduced~\cite{zhu2019deformable}, allowing the network to selectively emphasize or suppress specific regions:

\begin{equation}
Y(\mathbf{p}) = \sum_{\mathbf{k} \in \mathcal{R}} \alpha_k \cdot \mathbf{K}(\mathbf{k}) \cdot \mathbf{F}(\mathbf{p} + \mathbf{k} + \Delta \mathbf{p_k}).
\end{equation}

The modulation scalars are generated through an auxiliary convolutional layer followed by a sigmoid activation. This mechanism enables the network to focus more effectively on salient features, further enhancing the segmentation accuracy.

\subsection{Parameter-Free Attention: SimAM}

\begin{figure}[ht!]
    \centering
    \includegraphics[width=1\linewidth]{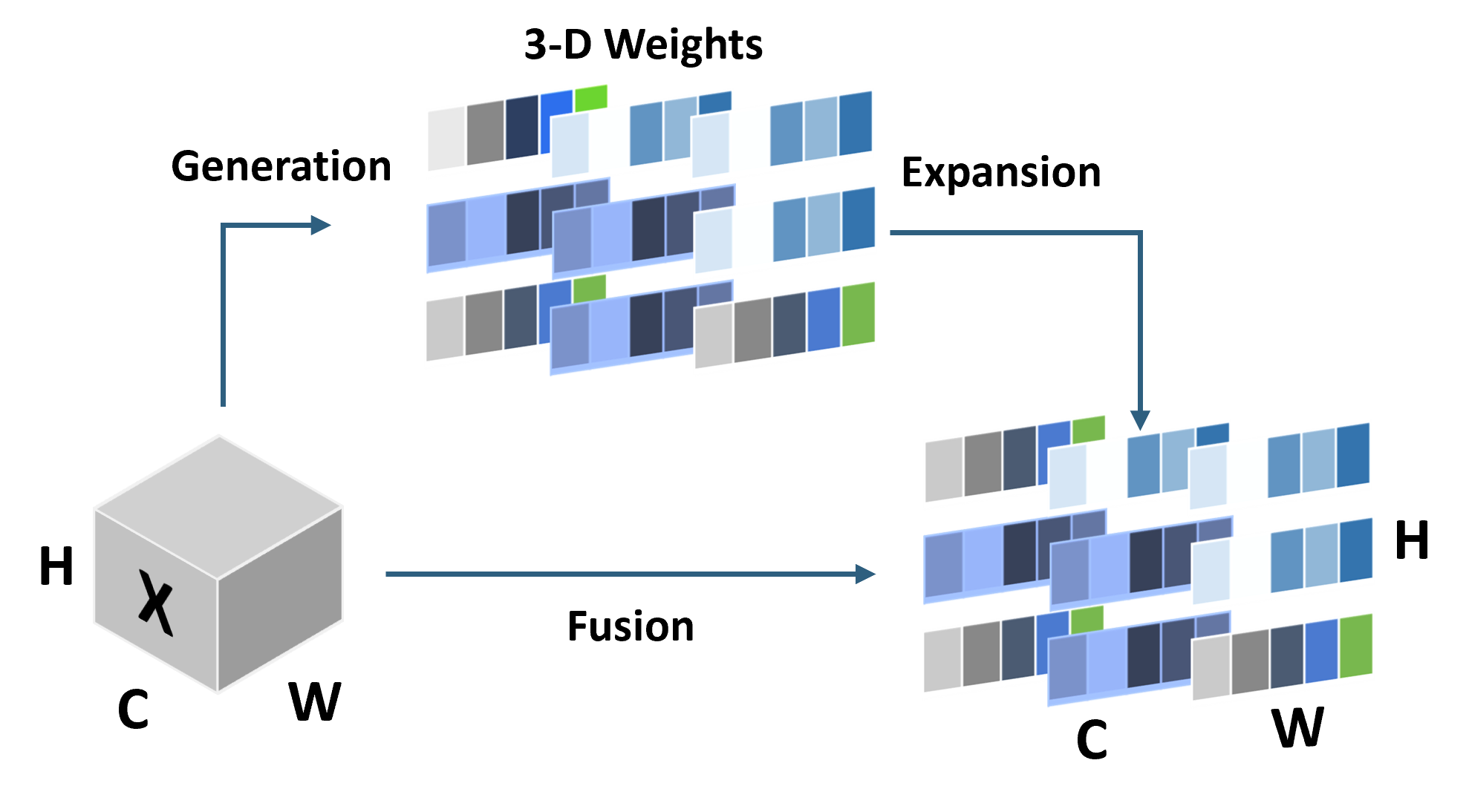}
    \caption{Schematic illustration of the SimAM attention mechanism. For each neuron, an energy-based evaluation is performed using surrounding spatial context, followed by an attention weighting operation. The resulting map highlights informative regions without introducing learnable parameters.}
    \label{fig:simAM_fig}
\end{figure}

SimAM is a parameter-free attention mechanism inspired by neuroscience theories, particularly the spatial suppression theory, which suggests that neurons with higher activation energies are less informative.

As illustrated in Figure~\ref{fig:simAM_fig}, SimAM performs attention modulation by evaluating each neuron's importance through an energy-based function, which considers both the neuron's deviation from the mean and the dispersion of surrounding neurons. This formulation allows SimAM to assign lower weights to high-energy (i.e., less informative) activations while highlighting those contributing meaningful context.

Given a feature map $\mathbf{X} \in \mathbb{R}^{C \times H \times W}$, SimAM computes an attention weight for each neuron based on the following energy function:

\begin{equation}
E_t = (x_t - \mu_t)^2 + \lambda \sum_{i \neq t} (x_i - \mu_t)^2,
\end{equation}

where $\mu_t$ is the mean of all neurons in the same channel excluding $x_t$, and $\lambda$ is a hyperparameter controlling the importance of surrounding neurons and it is set to be $\lambda= 1\times10^{-4}$.

The attention weight $a_t$ for each neuron is computed as:

\begin{equation}
a_t = \sigma\left(\frac{1}{E_t + \epsilon}\right),
\end{equation}

where $\sigma(\cdot)$ denotes the sigmoid activation, and $\epsilon$ is a small constant to avoid division by zero.

The refined output $\mathbf{X}'$ is obtained by element-wise multiplication of the original feature map and the attention weights:

\begin{equation}
\mathbf{X}' = \mathbf{X} \odot \mathbf{A},
\end{equation}

where $\odot$ denotes element-wise multiplication and $\mathbf{A}$ is the attention map composed of $a_t$ values across all spatial positions.

By applying SimAM, our model selectively amplifies informative features while suppressing less relevant ones, all without adding any additional parameters, thus achieving both enhanced accuracy and computational efficiency.

\subsection{Proposed Architecture}

\begin{figure*}[ht!]
    \centering
    \includegraphics[width=1\linewidth]{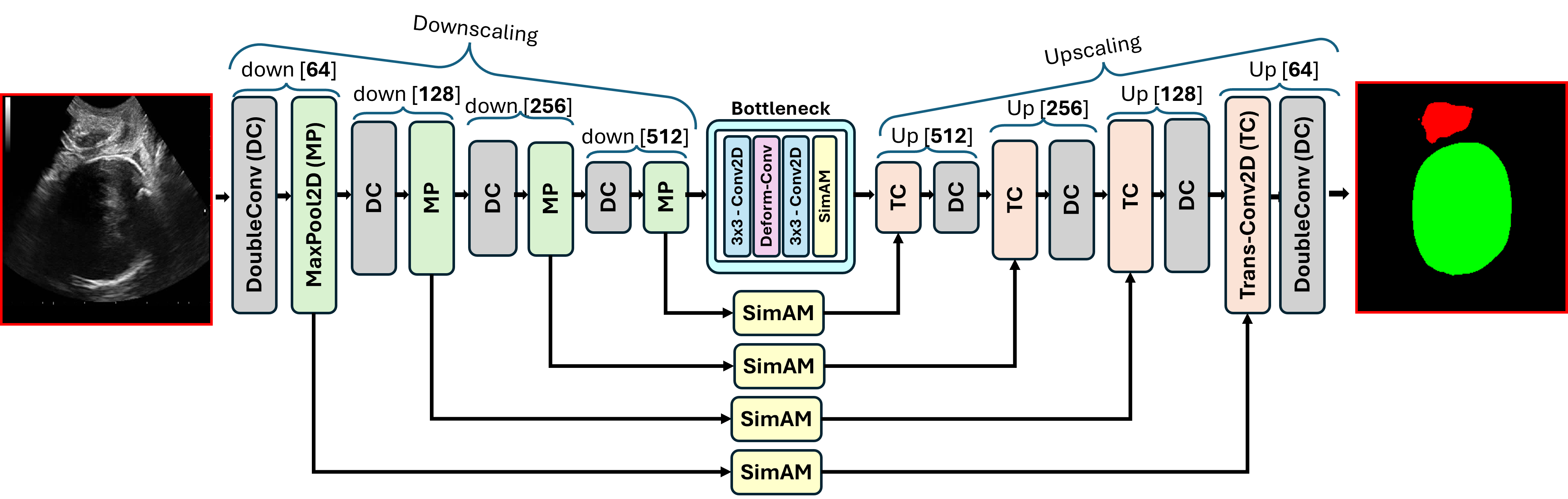}
    \caption{Overview of the proposed DAUNet architecture. The network is a lightweight variant of the UNet, incorporating key modifications in the bottleneck and skip connection paths. The bottleneck block is redesigned with a sequence of three operations: a $3 \times 3$ convolution for channel compression, a $3 \times 3$ Deformable Convolution V2 for adaptive spatial modeling, followed by another $3 \times 3$ convolution and a SimAM attention block. Additionally, the skip connections are augmented with SimAM modules to enhance feature fusion and emphasize informative activations without increasing parameter count.}
    \label{fig:network_arch}
\end{figure*}

The architecture of the proposed DAUNet is shown in Figure~\ref{fig:network_arch}. It retains the encoder–decoder structure of the classical UNet but introduces key improvements to enhance representational power and efficiency. In both the encoder and the decoder blocks, all convolution operations uses a $3 \times 3$ size filters. The bottleneck block is significantly redesigned by replacing the conventional convolution with a compound structure composed of four consecutive operations. First, a $3 \times 3$ convolution is used to compress the input channels to a quarter of the target output channels, and control computational cost. This is followed by a $3 \times 3$ Deformable Convolution V2 layer~\cite{dai2017deformable,zhu2019deformable}, which introduces spatial adaptability by learning dynamic offsets to the sampling grid, effectively capturing geometric variations in the input features. A second $3 \times 3$ convolution is then applied to project the features back to the original channel dimension. Finally, a SimAM module is appended to refine the output by emphasizing spatially informative activations based on an energy-based criterion, without introducing any learnable parameters.

In addition to the modified bottleneck, we also enhance all skip connection pathways by integrating SimAM modules before merging the encoder and decoder features. This modification helps suppress irrelevant or redundant activations and strengthens the transfer of semantically rich features across the network. These lightweight yet effective modifications improve segmentation accuracy and generalization, especially in challenging medical imaging scenarios, without significantly increasing the model’s parameter count or computational burden.

\section{Experimental Setup}\label{sec:experiments}
The experimental setup of the study is defined as follows:
\subsection{Training Setup}
The CNN models were trained using the Adam optimizer~\cite{adam}, with a fixed learning rate of $0.0001$. Training was performed on an NVIDIA RTX A4500 GPU, running each model for 100 epochs with a batch size of $16$. To maintain consistency during evaluation, all input images were resized to a fixed resolution of $256\times256$ pixels. 

The models were developed using the PyTorch framework~\cite{pytorch}, leveraging GPU acceleration through CUDA version 12.2. For the proposed approach, a hybrid loss function was employed, combining Dice loss with $0.5$ weighted Binary Cross-Entropy (BCE)~\cite{bce} to effectively guide the learning process.

We also used geometric transformation-based data augmentation methods such as adaptive zooming, rotation, and random flipping.  By boosting the variety of rare class appearances, these augmentations not only add diversity to the training data to avoid overfitting but also aid in addressing class imbalance.  

\subsection{Dataset}
The proposed model is evaluated on two datasets:
\begin{itemize}
    \item \textbf{Pubic Symphysis and Fetal Head Detection (FH-PS-AoP):} 
    The FH-PS-AoP dataset~\cite{PSFH} comes from the Pubic Symphysis–Fetal Head Segmentation Challenge, which deals with transperineal ultrasound imaging. It includes 2D B-mode ultrasound images from different regions of China, gathered with standard ultrasound machines. The dataset contains information about pregnant women between 18 and 46 years old. Annotation was conducted using Pair software and was performed by a group of seven annotators composed of two senior specialists and five undergraduates. The entire dataset contains 4,000 images for training and 700 for testing. For model training, the 4,000 training samples were divided into an $80:20$ ratio for training and validation to monitor potential overfitting. Lastly, the 700 test samples were used for performance evaluation.

    \item \textbf{Pulmonary Embolism Detection (FUMPE Dataset):} The FUMPE dataset (Ferdowsi University of Mashhad's Pulmonary Embolism dataset) ~\cite{fumpe} contains the three-dimensional computed tomography angiography (CTA) scans of 35 patients, amounting to 8792 image slices. Two radiologists manually reviewed all the CTA scans using a semi-automated segmentation method to derive reference labels. In total, the dataset consists of 3,438 PE (pulmonary embolism) annotated regions. Notably, approximately 67\% of these regions occur in the peripheral pulmonary arteries, which positions this dataset to be highly advantageous for prototyping and benchmarking sophisticated CAD systems. After a comprehensive review, we completed the collection of CTPA images for 35 patients, totaling 8792 CTPA images, along with updated annotations by the doctors. For performance evaluation, the dataset was divided into an $80:20$ ratio for training and validation. And performance metric were reported on unseen validation samples as suggested in~\cite{scunetpp}.
\end{itemize}

\new{Further details regarding data splits, evaluation protocols, and fairness of comparison across datasets are provided in Section~\ref{sec:data_fairness}.}

\subsection{Evaluation Metrics}

To quantitatively assess the segmentation performance of DAUNet and compare it against other state-of-the-art models, we employ three widely adopted metrics in medical image segmentation: Dice Similarity Coefficient (DSC)~\cite{DSC}, 95th percentile Hausdorff Distance (HD95)~\cite{HD}, and Average Symmetric Surface Distance (ASD)~\cite{ASD}. These metrics provide a comprehensive evaluation across both region-based and boundary-based performance criteria. Lastly, we use parameter count as a metric for model complexity.

\subsubsection{Dice Similarity Coefficient (DSC)}

The Dice Similarity Coefficient measures the overlap between the predicted segmentation mask \( P \) and the ground truth mask \( G \). It is defined as:

\begin{equation}
\text{DSC} = \frac{2 |P \cap G|}{|P| + |G|},
\end{equation}

where \( |\cdot| \) denotes the cardinality of a set. DSC ranges from 0 to 1, where a higher value indicates better agreement.

\subsubsection{Hausdorff Distance (HD95)}

Hausdorff Distance quantifies the maximum boundary deviation between the predicted and ground truth masks. We use the 95th percentile version for robustness to outliers:

\begin{equation}
\begin{split}
\text{HD}_{95}(P, G) = \max \Big\{ \sup_{p \in \partial P} \inf_{g \in \partial G} \|p - g\|, \\
\sup_{g \in \partial G} \inf_{p \in \partial P} \|g - p\| \Big\}_{95\%},
\end{split}
\end{equation}

where \( \partial P \) and \( \partial G \) are the boundaries of the predicted and ground truth masks. Lower HD95 values indicate better boundary alignment.

\subsubsection{Average Symmetric Surface Distance (ASD)}

ASD measures the average bidirectional distance between the boundaries of prediction and ground truth:

\begin{equation}
\begin{split}
\text{ASD}(P, G) = \frac{1}{|\partial P| + |\partial G|} \Big( \sum_{p \in \partial P} \min_{g \in \partial G} \|p - g\| \\
+ \sum_{g \in \partial G} \min_{p \in \partial P} \|g - p\| \Big).
\end{split}
\end{equation}

ASD provides an interpretable measure of the mean boundary error, with lower values indicating better segmentation precision.

\subsubsection{Model Complexity (Parameter Count)}

In addition to accuracy-based metrics, we also report the number of trainable parameters as an indicator of model complexity. Lower parameter counts are desirable in clinical scenarios involving real-time processing or deployment on edge devices. 

\section{Experimental Results}\label{sec:results}

We conduct extensive experiments on two challenging medical image segmentation tasks, including pubic symphysis and fetal head segmentation from ultrasound (FH-PS-AoP dataset), and pulmonary embolism detection from CT angiography scans (FUMPE dataset), to evaluate the performance of the proposed DAUNet model.

\subsection{Comparison with State-of-the-Art Methods}

\subsubsection{\new{Experimental Protocol and Fairness of Comparison}}
\label{sec:data_fairness}

\new{To ensure a fair and reproducible comparison, different strategies were adopted depending on dataset availability and evaluation standards.}

\new{For the PSFHS (FH-PS-AoP) dataset, evaluation was conducted on the official challenge Testing Set 2, consisting of 700 images from 545 different subjects, which serves as a fully patient-independent held-out test set. This evaluation protocol follows the challenge guidelines and ensures fair comparison with prior work \cite{bai2025psfhs}. Results for a subset of methods are taken from the \cite{bai2025psfhs,zhou2025segment}, where standardized preprocessing, data splits, and evaluation metrics are enforced. For methods that were not part of the official challenge but reported results on the same dataset, their original publications are referenced. In cases where methods were neither included in the challenge nor had publicly reported results, we reimplemented them from scratch following the preprocessing, augmentation, and evaluation procedures specified in the challenge documentation \cite{bai2025psfhs}.}

\new{For the FUMPE dataset, the data were split at the patient level, with 28 patients (80\%) used for training and the remaining 7 patients (20\%) reserved for validation. All reported results are obtained on unseen patients, thus preventing data leakage and ensuring robust generalization assessment. Results for a subset of methods are taken from \cite{scunetpp}. While all other methods were implemented and trained from scratch using identical preprocessing, augmentation, and evaluation settings. Training and validation splits strictly follow the protocol described in the original SCUNet++\cite{scunetpp} study, which serves as the primary reference for dataset usage. This ensures that all methods are evaluated under the same conditions.}

\new{In addition to quantitative evaluation, several representative baselines, including nnUNet, TransUNet, TransAttUNet, SCUNet++, and FAT-Net, were implemented under identical inference settings for qualitative comparison. Visual results are presented in Figures~\ref{fig:GC} and~\ref{fig:Fsam}, illustrating the relative segmentation quality across different methods. To ensure the reproducibility of proposed method code and weights of the pretrained model are provided at \url{https://github.com/Shujaat123/DAUNet}.}

\subsubsection{Pubic Symphysis and Fetal Head}
\begin{table*}[!ht]
\centering
\caption{\new{Quantitative comparison of DAUNet and state-of-the-art models on the FH-PS-AoP dataset. Performance is reported using Dice Similarity Coefficient (DSC $\uparrow$), 95th percentile Hausdorff Distance (HD95 $\downarrow$), and Average Symmetric Surface Distance (ASD $\downarrow$).}}
\label{tab:table5_segmentation}
\resizebox{\linewidth}{!}{
\begin{tabular}{c|ccc|ccc|ccc|c}
\hline \hline
\multirow{2}{*}{\bf Model} 
& \multicolumn{3}{c|}{\bf DSC $\uparrow$} 
& \multicolumn{3}{c|}{\bf HD95 $\downarrow$} 
& \multicolumn{3}{c|}{\bf ASD $\downarrow$} 
& \bf Param (M) \\ \cline{2-10}

& \bf FH & \bf PS & \bf PSFH 
& \bf FH & \bf PS & \bf PSFH 
& \bf FH & \bf PS & \bf PSFH 
&  \\ \hline \hline

UNet~\cite{unet} 
& $91.25 \pm 0.07$ & $82.80 \pm 0.17$ & $90.55 \pm 0.06$
& $17.25 \pm 0.18$ & $9.34 \pm 0.91$ & $16.99 \pm 0.16$
& $5.55 \pm 0.77$ & $3.15 \pm 0.31$ & $4.86 \pm 0.47$
& 31.03 \\ \hline

AttUNet~\cite{attUNet}
& $91.10\pm0.42$ & $79.25\pm2.35$ & $90.08\pm0.36$
& $17.49 \pm 0.55$ & $13.86 \pm 1.40$ & $20.52 \pm 0.67$
& $5.13 \pm 0.23$ & $4.93 \pm 0.90$ & $4.76 \pm 0.17$
& 35.6 \\ \hline

nnUNetv2~\cite{isensee2021nnu}
& $92.50 \pm 0.17$ & $82.30 \pm 0.73$ & $91.50 \pm 0.18$
& $13.71 \pm 0.33$ & $10.79 \pm 0.77$ & $15.48 \pm 0.25$
& $4.29 \pm 0.11$ & $3.75 \pm 0.27$ & $\mathbf{3.00 \pm 0.09}$
& 92.5 \\ \hline

TransUNet~\cite{transunet}
& $\mathbf{93.02 \pm 0.05}$ & $81.66 \pm 1.25$ & $91.95 \pm 0.13$
& $13.25 \pm 0.44$ & $11.36 \pm 0.56$ & $15.78 \pm 0.45$
& $\mathbf{3.96 \pm 0.05}$ & $3.59 \pm 0.18$ & $3.76 \pm 0.08$
& 105.3 \\ \hline

SwinUNet~\cite{swinunet}
& $92.06 \pm 0.10$ & $82.84 \pm 0.19$ & $91.15 \pm 0.09$
& $13.96 \pm 0.07$ & $11.06 \pm 0.15$ & $16.14 \pm 0.09$
& $4.38 \pm 0.05$ & $3.45 \pm 0.08$ & $4.05 \pm 0.04$
& 27.2 \\ \hline

DSEUNet~\cite{DSUNet}
& $92.44 \pm 0.34$ & $84.86 \pm 0.23$ & $91.75 \pm 0.31$
& $16.47 \pm 0.90$ & $12.67 \pm 1.83$ & $18.01 \pm 0.79$
& $4.47 \pm 0.22$ & $3.35 \pm 0.12$ & $3.98 \pm 0.17$
& 62.9 \\ \hline

UPerNet~\cite{upernet}
& $92.24 \pm 0.15$ & $80.67 \pm 0.97$ & $91.09 \pm 0.19$
& $14.37 \pm 0.26$ & $12.60 \pm 0.91$ & $17.12 \pm 0.50$
& $4.43 \pm 0.08$ & $3.80 \pm 0.29$ & $4.20 \pm 0.08$
& 31.3 \\ \hline

MissFormer~\cite{MFormer}
& $91.58 \pm 0.95$ & $80.61 \pm 2.33$ & $90.54 \pm 01.01$
& $15.78 \pm 1.11$ & $13.65 \pm 0.83$ & $18.67 \pm 1.31$
& $4.82 \pm 0.62$ & $3.97 \pm 0.57$ & $4.45 \pm 0.50$
& 42.3 \\ \hline

H2Former~\cite{HFormer}
& $92.64 \pm 0.29$ & $\mathbf{85.32 \pm 0.45}$ & $91.94 \pm 0.26$
& $14.06 \pm 0.22$ & $11.07 \pm 0.64$ & $16.25 \pm 0.14$
& $4.20 \pm 0.15$ & $2.97 \pm 0.14$ & $3.78 \pm 0.11$
& 33.9 \\ \hline

FAT-Net~\cite{fatnet}
& $92.94 \pm 0.15$ & $80.78 \pm 1.45$ & $91.90 \pm 0.21$
& $14.13 \pm 0.24$ & $12.45 \pm 0.83$ & $16.71 \pm 0.41$
& $3.99 \pm 0.07$ & $4.58 \pm 0.57$ & $3.80 \pm 0.10$
& 28.8 \\ \hline

CE-Net~\cite{CENet}
& $92.68 \pm 0.44$ & $84.90 \pm 0.25$ & $\mathbf{91.96 \pm 0.38}$
& $14.57 \pm 0.89$ & $12.55 \pm 0.89$ & $16.21 \pm 0.92$
& $4.16 \pm 0.27$ & $3.33 \pm 0.28$ & $3.77 \pm 0.19$
& 29.0 \\ \hline

MedSAM~\cite{Medsam}
& $90.59 \pm 0.40$ & $83.85 \pm 1.08$ & $89.95 \pm 0.40$
& $16.37 \pm 0.98$ & $9.52 \pm 0.67$ & $17.19 \pm 0.83$
& $5.26 \pm 0.26$ & $\mathbf{2.79 \pm 0.17}$ & $4.54 \pm 0.20$
& 93.7 \\ \hline

Proposed
& $92.87 \pm 0.06$ & $85.31 \pm 0.13$ & $91.95 \pm 0.05$
& $\mathbf{12.76 \pm 0.13}$ & $\mathbf{8.01 \pm 0.09}$ & $\mathbf{12.83 \pm 0.12}$
& $4.39 \pm 0.04$ & $2.95 \pm 0.03$ & $3.93 \pm 0.03$
& $\mathbf{21.07}$ \\ \hline \hline

\end{tabular}
}
\end{table*}
\new{Table~\ref{tab:table5_segmentation} reports a comprehensive quantitative comparison between the proposed DAUNet and a range of state-of-the-art segmentation models on Pubic Symphysis and Fetal Head dataset. To improve statistical rigor, all results are reported as mean $\pm$ standard deviation across the test set for Dice Similarity Coefficient (DSC), 95th percentile Hausdorff Distance (HD95), and Average Symmetric Surface Distance (ASD).}

\new{In terms of overlap accuracy, DAUNet achieves competitive DSC values across FH, PS, and PSFH, ranking among the top-performing methods while maintaining a substantially smaller model size. Although some competing approaches achieve marginally higher DSC on individual classes, these gains are typically associated with significantly higher parameter counts.}

\new{Boundary-based metrics further emphasize the strengths of the proposed approach. DAUNet consistently achieves the lowest HD95 across all classes, indicating improved robustness to outliers and more accurate contour delineation. For ASD, DAUNet remains competitive across all structures, achieving low and stable values without increasing computational complexity.}

\new{To assess the reliability of these differences, we conducted statistical significance analysis using 700 independent test samples. Paired hypothesis testing against representative competitive baselines confirms that DAUNet achieves statistically significant improvements in HD95 and ASD across all classes ($p < 0.05$), while DSC differences are statistically significant for PS and PSFH and statistically comparable for FH. The large sample size yields narrow confidence intervals, further supporting the robustness of the reported results.}

\new{Overall, DAUNet offers a favorable balance between segmentation accuracy, boundary precision, and computational efficiency, making it well suited for real-time and resource-constrained clinical applications.}

\subsubsection{FUMPE}

\begin{table}[!h]
\centering
\caption{Performance comparison of DAUNet and competing models on the FUMPE dataset. DAUNet achieves the best segmentation accuracy (DSC $\uparrow$) and lowest boundary error (HD95 $\downarrow$) with fewer parameters, demonstrating superior efficiency and robustness.}

\begin{tabular}{c|c|c|c}

\hline \hline
\bf Methods           & \bf DSC $\uparrow$          & \bf HD95  $\downarrow$        & \bf Param (M) $\downarrow$\\ \hline \hline
UNet~\cite{unet}              & 77.91          & 6.86          & 31.03                                                     \\ \hline
Unet ++~\cite{unet++}           & 77.16          & 5.80          & 34.96                                                     \\ \hline
SwinUNet~\cite{swinunet}          & 60.80          & 20.20         & 25.91                                                     \\ \hline
ResD-UNet~\cite{resunet}         & 76.48          & 22.25         & 50.83                                                     \\ \hline
TransAttUNet~\cite{trans}         & 82.82          & 3.2         & 25.96                                                    \\ \hline
SCUNet++~\cite{scunetpp}          & 83.47          & 3.83          & 60.11                                                     \\ \hline
FAT-Net~\cite{fatnet}          & 84.44          & 3.67          & 30.00                                                     \\ \hline
DAUNet (Proposed)& \textbf{88.80} & \textbf{2.57} & \textbf{21.07}                                            \\ \hline \hline
\end{tabular}
\label{tab:fumpe_results}
\end{table}

Table~\ref{tab:fumpe_results} presents a quantitative comparison between the proposed DAUNet and several representative state-of-the-art segmentation models, including CNN-based, transformer-based, and hybrid architectures. The comparison is conducted on the FUMPE dataset using Dice Similarity Coefficient (DSC) and 95th percentile Hausdorff Distance (HD95), together with model parameter count as a measure of computational efficiency.

As shown in Table~\ref{tab:fumpe_results}, the proposed DAUNet achieves the best overall segmentation performance, attaining a Dice score of \textbf{88.80\%} and the lowest boundary error with an HD95 of \textbf{2.57}. These results indicate superior overlap accuracy and more precise boundary delineation compared to all competing methods. In addition, DAUNet maintains a significantly lower model complexity, requiring only \textbf{21.07M} parameters, which is substantially fewer than several strong baselines.

Notably, models with considerably higher parameter counts, such as SCUNet++ (60.11M), do not achieve comparable segmentation accuracy or boundary robustness on this dataset. Transformer-based and hybrid models, including TransAttUNet and FAT-Net, demonstrate improved performance over classical UNet variants; however, they remain inferior to DAUNet in both DSC and HD95, while requiring higher computational cost.

Overall, these results highlight the effectiveness of the proposed lightweight architecture for pulmonary embolism segmentation. DAUNet consistently delivers superior accuracy and robustness while maintaining a compact model size, making it well suited for deployment in real-time and resource-constrained clinical environments.

\subsection{Qualitative Analysis}

\begin{figure*}[t]
\centering
\resizebox{\textwidth}{!}{
\begin{tabular}{c@{ }c@{ }c@{ }c@{ }c@{ }c@{ }c}
  \includegraphics[width=0.18\textwidth]{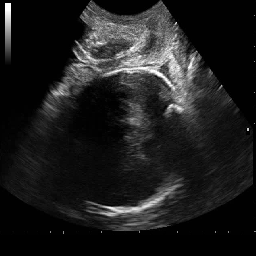}&  
 \includegraphics[width=0.18\textwidth]{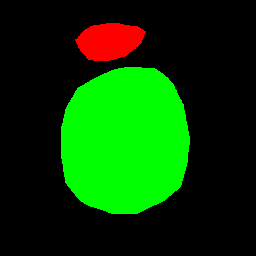}  &
 \includegraphics[width=0.18\textwidth]{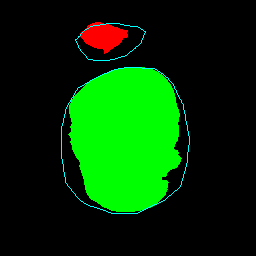}    &
  \includegraphics[width=0.18\textwidth ]{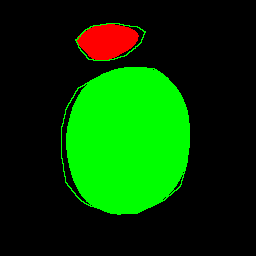} &
 \includegraphics[width=0.18\textwidth ]{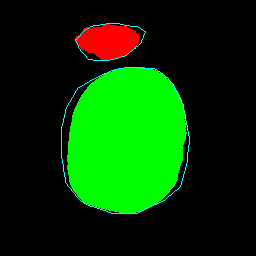} & 
 \includegraphics[width=0.18\textwidth ]{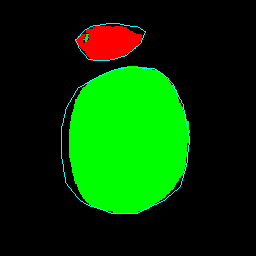} &  \includegraphics[width=0.18\textwidth ]{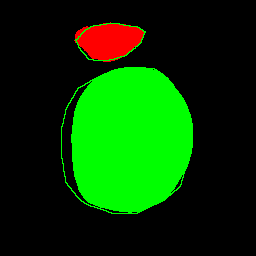} \\
  \includegraphics[width=0.18\textwidth]{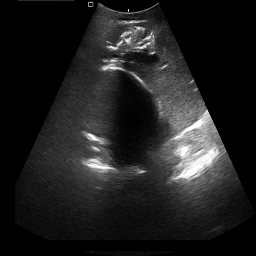}&
 \includegraphics[width=0.18\textwidth]{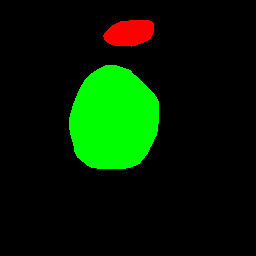} &
  \includegraphics[width=0.18\textwidth]{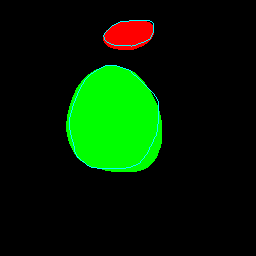} &
  \includegraphics[width=0.18\textwidth]{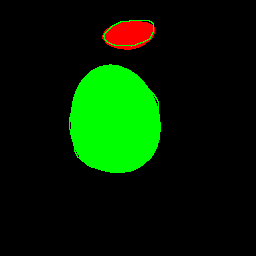} & 
  \includegraphics[width=0.18\textwidth]{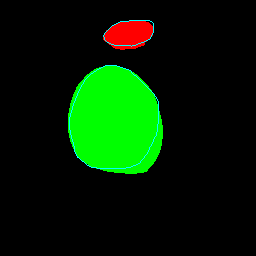}
  &  \includegraphics[width=0.18\textwidth ]{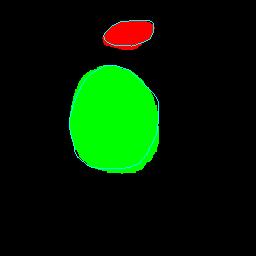} &
  \includegraphics[width=0.18\textwidth]{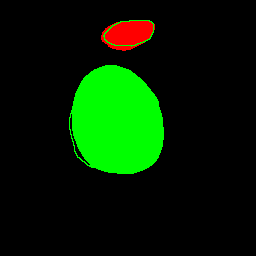} \\
  
  \includegraphics[width=0.18\textwidth]{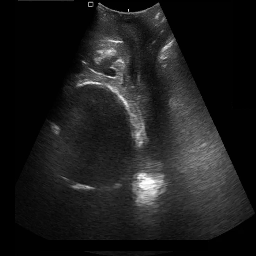}&
 \includegraphics[width=0.18\textwidth]{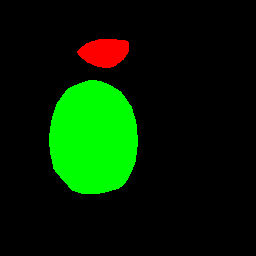} &
  \includegraphics[width=0.18\textwidth]{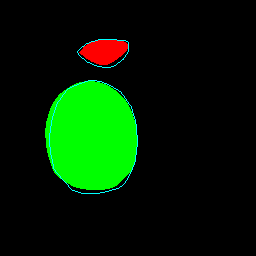} &
  \includegraphics[width=0.18\textwidth]{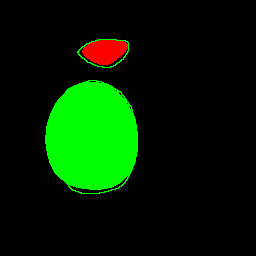} 
  & 
  \includegraphics[width=0.18\textwidth ]{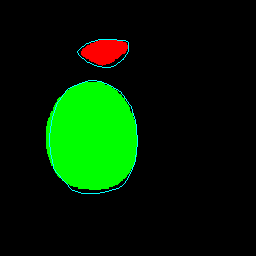} & 
    \includegraphics[width=0.18\textwidth]{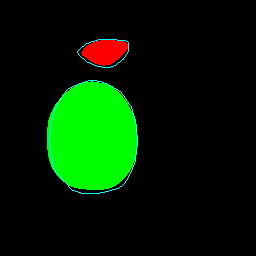} 
  &\includegraphics[width=0.18\textwidth]{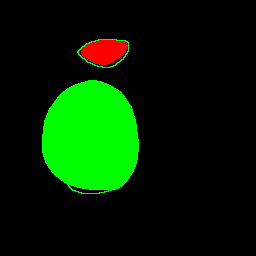} \\
  Input & Ground-Truth & UNet & \new{nnU-Netv2} & TransUNet & FAT-Net & DAUNet (Proposed)  \\
\end{tabular} }
\caption{ Segmentation results of different models on FH-PS-AoP dataset: (a) Input image, (b), The ground truth mask, (c) UNet, (d) \new{nnU-Netv2}, (e) TransUNet (f) FAT-Net , (g) DAUNet (Proposed). The contours around the prediction masks are the ground truth mask contours. 
 }
\label{fig:GC}
\end{figure*}

\begin{figure*}[t]
\centering
\resizebox{\textwidth}{!}{
\begin{tabular}{c@{ }c@{ }c@{ }c@{ }c@{ }c@{ }c}
  \includegraphics[width=0.18\textwidth ]{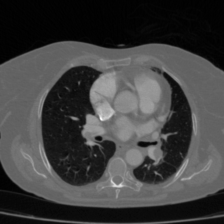}&  
 \includegraphics[width=0.18\textwidth]{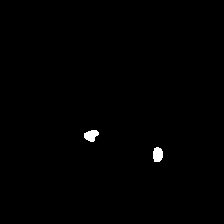}  &
 \includegraphics[width=0.18\textwidth]{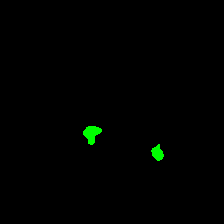}    &
 \includegraphics[width=0.18\textwidth ]{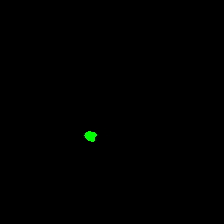} &  
 \includegraphics[width=0.18\textwidth ]{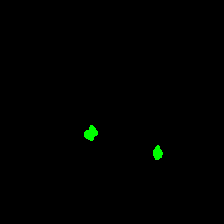} &
 \includegraphics[width=0.18\textwidth ]{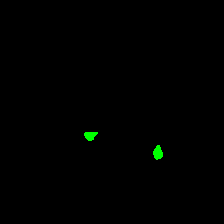} &
 \includegraphics[width=0.18\textwidth ]{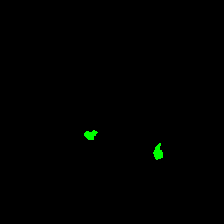} \\
  \includegraphics[width=0.18\textwidth]{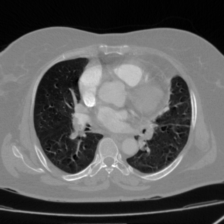}&
 \includegraphics[width=0.18\textwidth]{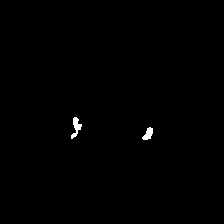} &
  \includegraphics[width=0.18\textwidth]{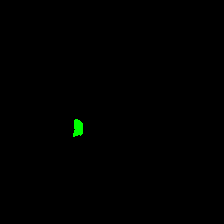} &
  \includegraphics[width=0.18\textwidth]{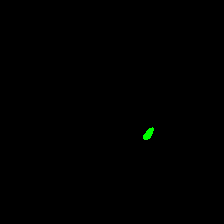} &
   \includegraphics[width=0.18\textwidth ]{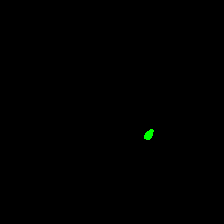} &
  \includegraphics[width=0.18\textwidth ]{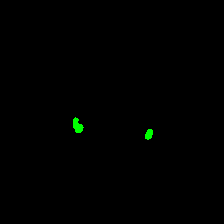} &
  \includegraphics[width=0.18\textwidth ]{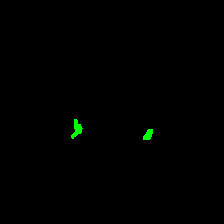} \\
  \includegraphics[width=0.18\textwidth]{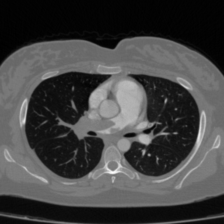}&
 \includegraphics[width=0.18\textwidth]{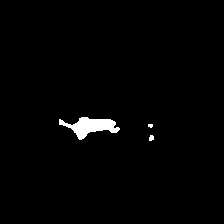} &
  \includegraphics[width=0.18\textwidth]{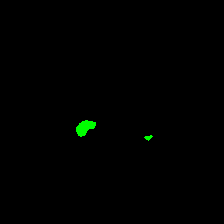} &
  \includegraphics[width=0.18\textwidth]{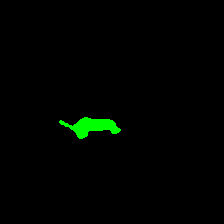} &
  \includegraphics[width=0.18\textwidth ]{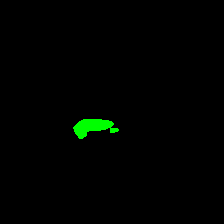} &
  \includegraphics[width=0.18\textwidth ]{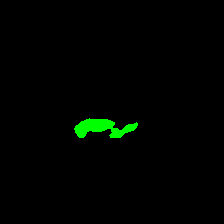} & \includegraphics[width=0.18\textwidth]{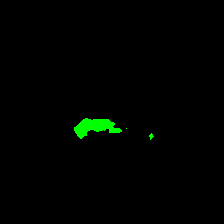} \\
  Input & Ground-Truth & UNet & SCUNet++&TransAttUNet &FAT-Net& DAUNet (Proposed)  \\
\end{tabular} }
\caption{ Segmentation results of different models on FUMPE dataset: (a) Input image, (b), The ground truth mask, (c) UNet, (d) SCUNet++, (e) TransAttUNet (f) FAT-Net , (g) DAUNet (Proposed).
 }
\label{fig:Fsam}
\end{figure*}

Figure~\ref{fig:GC} and Figure~\ref{fig:Fsam} provide visual comparisons of segmentation masks generated by DAUNet and other competing methods on representative samples from both datasets. As illustrated in Figure~\ref{fig:GC}, DAUNet produces more accurate and smoother boundary delineations, closely aligning with the ground truth annotations, even in challenging low-contrast regions or under partial occlusion. A similar performance can be observed in Figure~\ref{fig:Fsam}, where proposed model accurately identify the pulmonary embolism regions. It is worth nothing that the proposed model achieve best performance in two different imaging modalities, consisting of different sized regions of interest and number of classes demonstrating a good generalization.

\subsection{\new{Computational and Parameter Efficiency Analysis}}

\begin{figure}[ht]
    \centering
    \includegraphics[width=1\linewidth]{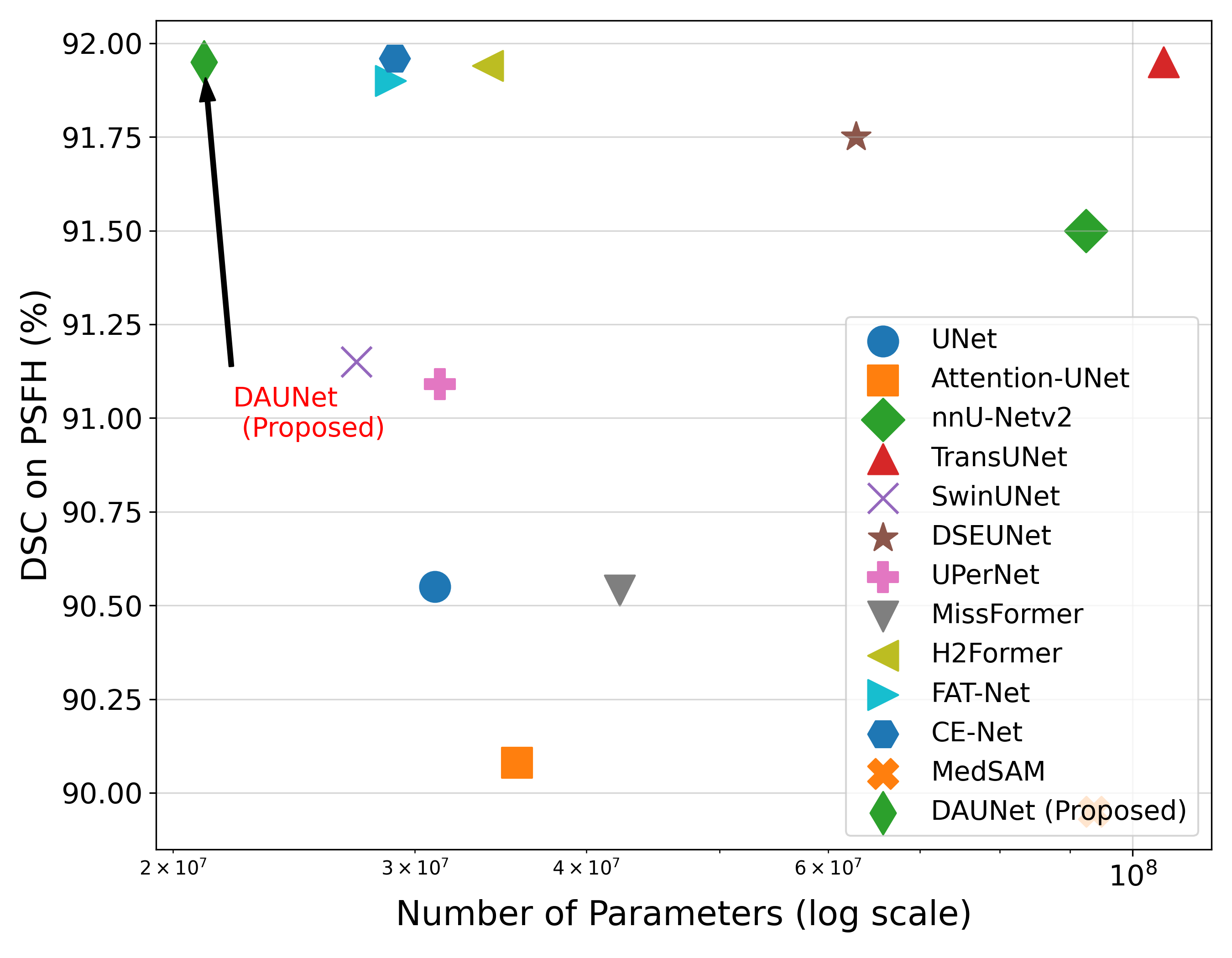}
    \caption{\new{Trade-off between model complexity and segmentation performance on the PSFH subset of the FH-PS-AoP dataset. The x-axis represents the number of trainable parameters (log scale), while the y-axis shows the Dice Similarity Coefficient (DSC). DAUNet (Proposed) achieves the highest DSC with the fewest parameters among all methods, demonstrating its superior efficiency and accuracy.}}
    \label{fig:pvd}
\end{figure}

To highlight the efficiency of DAUNet, Figure~\ref{fig:pvd} illustrates the trade-off between model complexity (in terms of parameter count) and segmentation performance (DSC) on FH-PS-AoP dataset. While most transformer-based and heavily parameterized CNN models achieve high accuracy at the cost of computational load, DAUNet attains state-of-the-art performance with significantly fewer parameters. 
\new{In terms of inference time, proposed model demonstrates its suitability for real-time applications, achieving an inference speed of approximately 144.3 frames per second (FPS), which corresponds to an average inference latency of 6.93 ms per frame on an NVIDIA RTX A4500 GPU, with a computational cost of approximately 45.5 GFLOPs per inference forward pass.}
 This favorable balance makes DAUNet highly suitable for real-time deployment in clinical workflows and on resource-constrained platforms such as edge devices and mobile ultrasound systems \cite{khan2021contrast}.

\subsection{Robustness to Missing Context}

\begin{figure*}[t]
\centering
\resizebox{\textwidth}{!}{
\begin{tabular}{c@{}c@{ }c@{ }c@{ }c@{ }c}
  \rotatebox{90}{\hspace{10mm} \bf Input} &
  \includegraphics[width=0.18\textwidth]{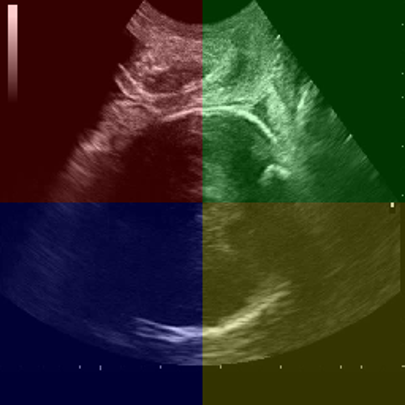} &  
  \includegraphics[width=0.18\textwidth]{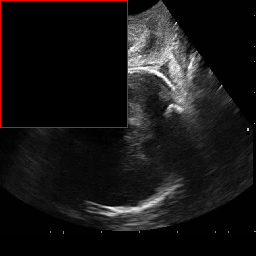} &
  \includegraphics[width=0.18\textwidth]{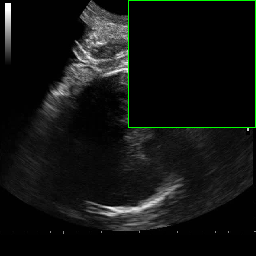} &
  \includegraphics[width=0.18\textwidth]{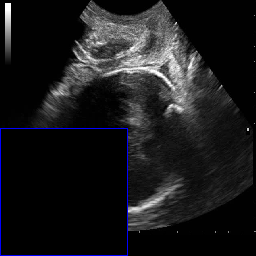} &
  \includegraphics[width=0.18\textwidth]{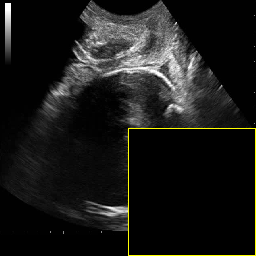} \\
  
  \rotatebox{90}{\hspace{6mm} \bf UNet (Offsets)} &
  \includegraphics[width=0.18\textwidth]{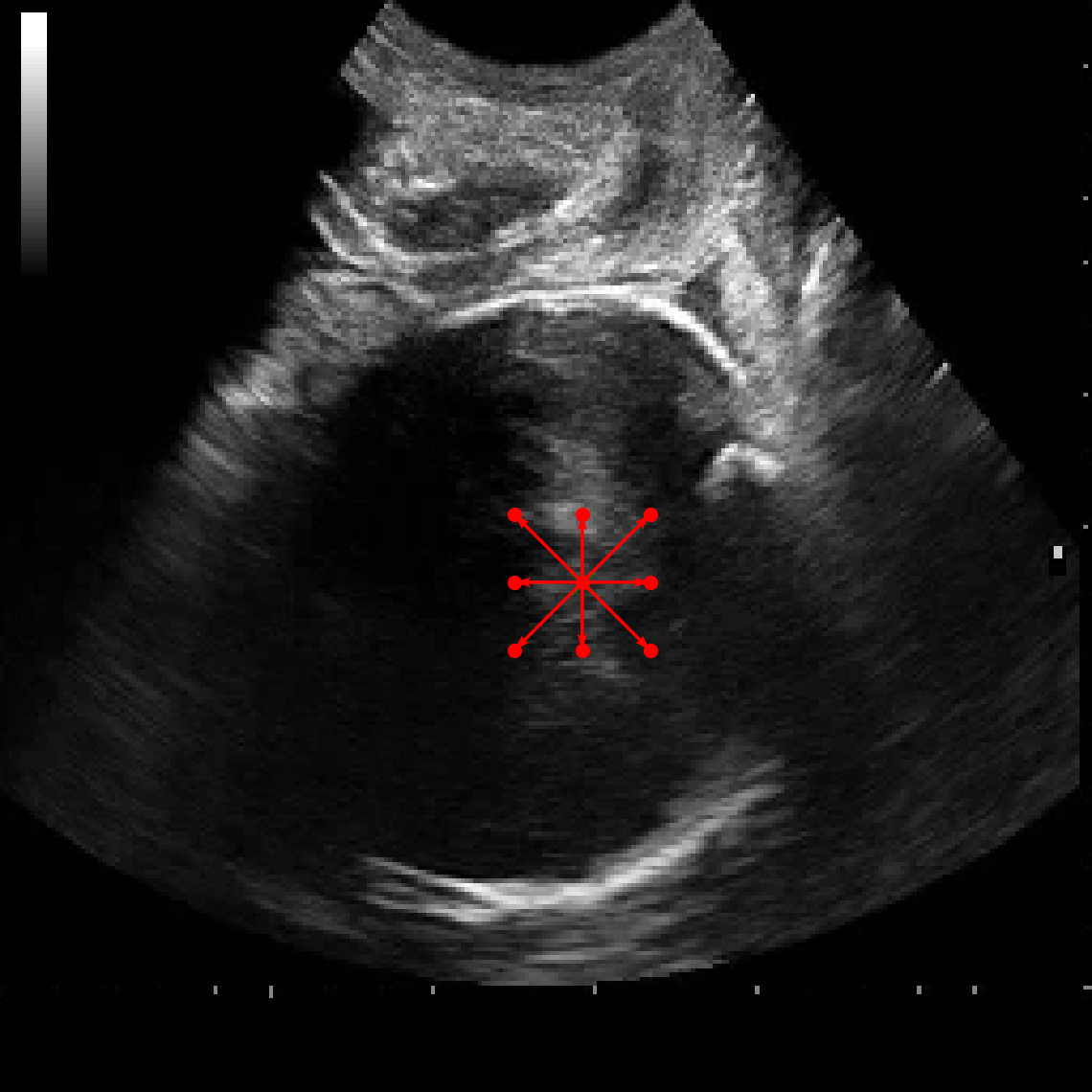} &
  \includegraphics[width=0.18\textwidth]{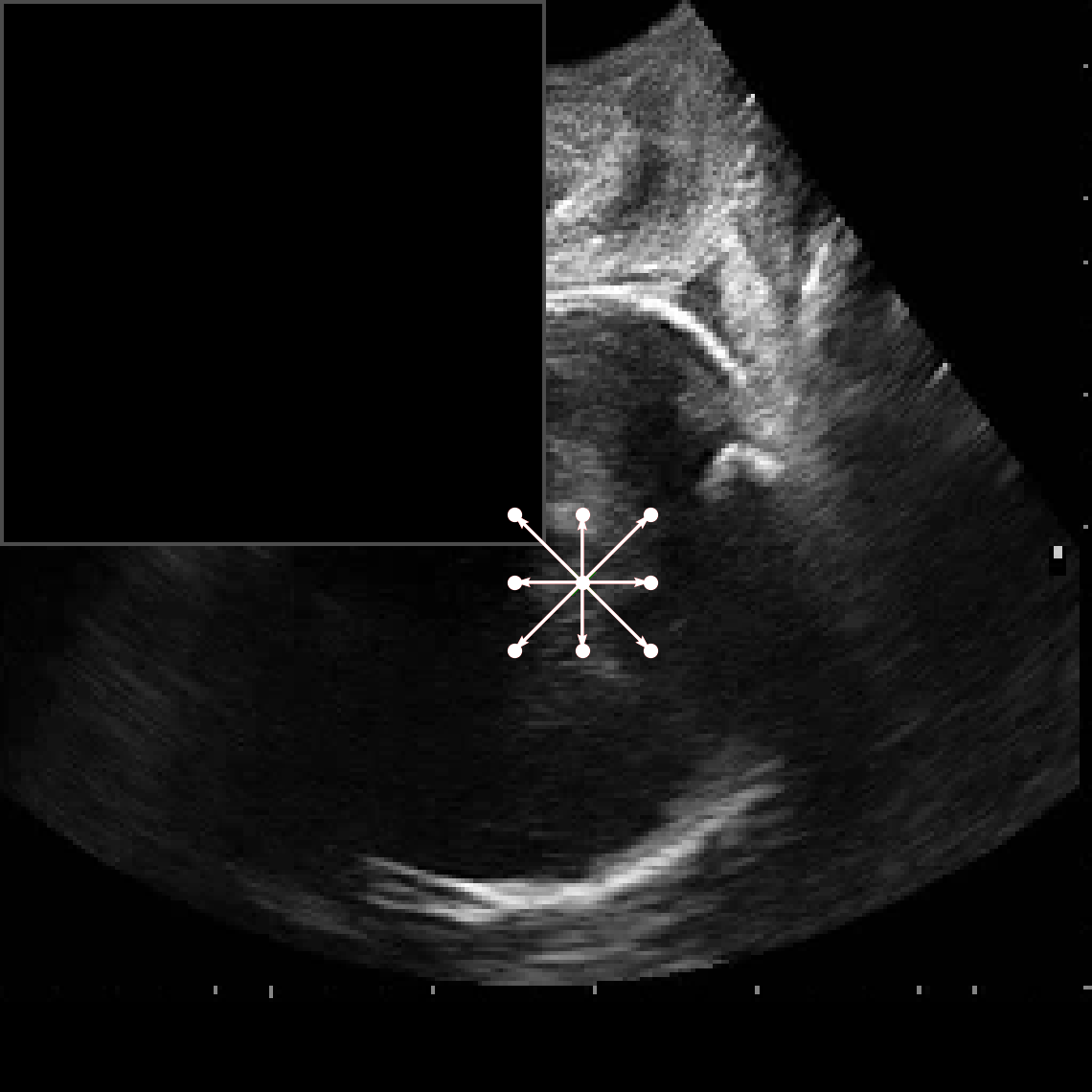} &
  \includegraphics[width=0.18\textwidth]{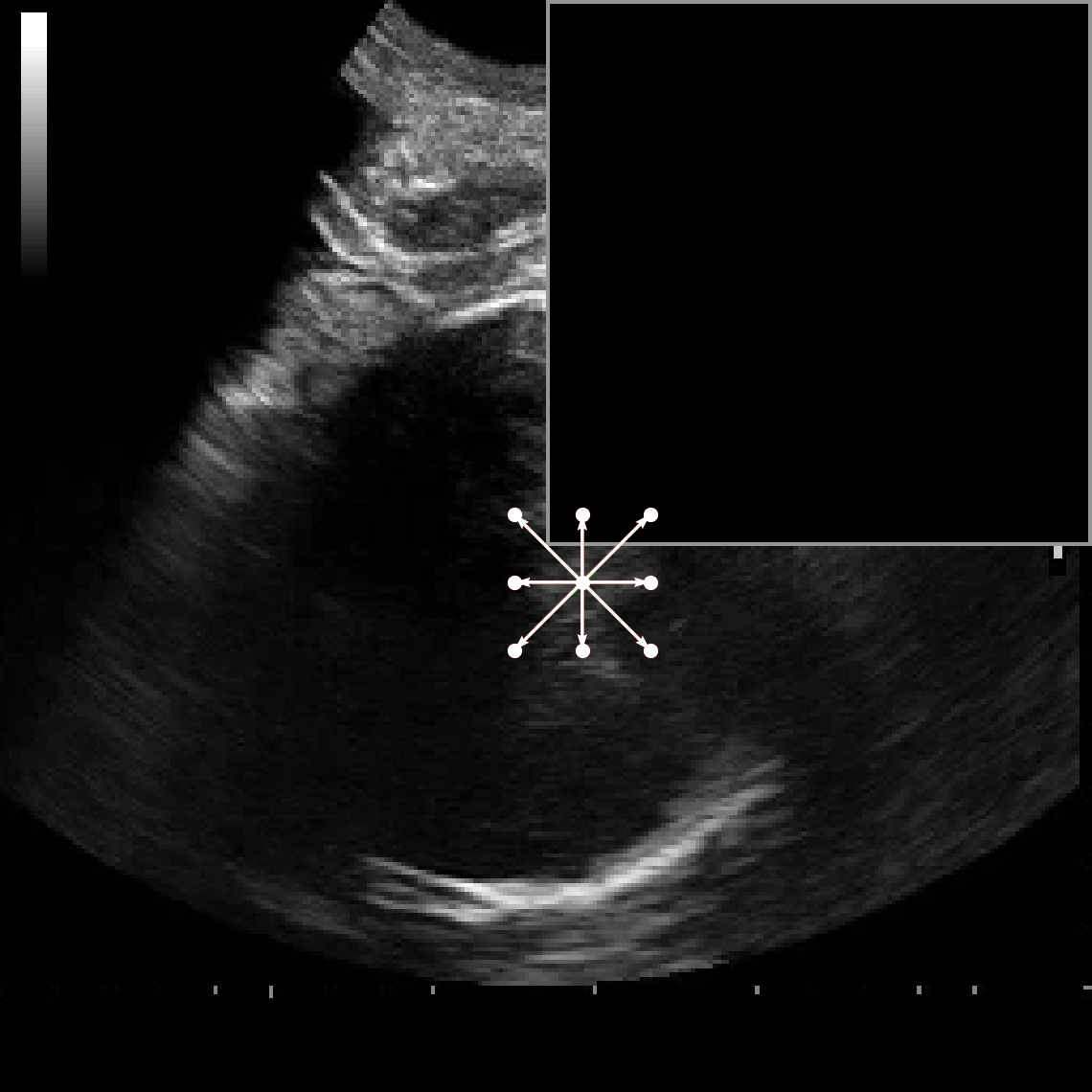} &
  \includegraphics[width=0.18\textwidth]{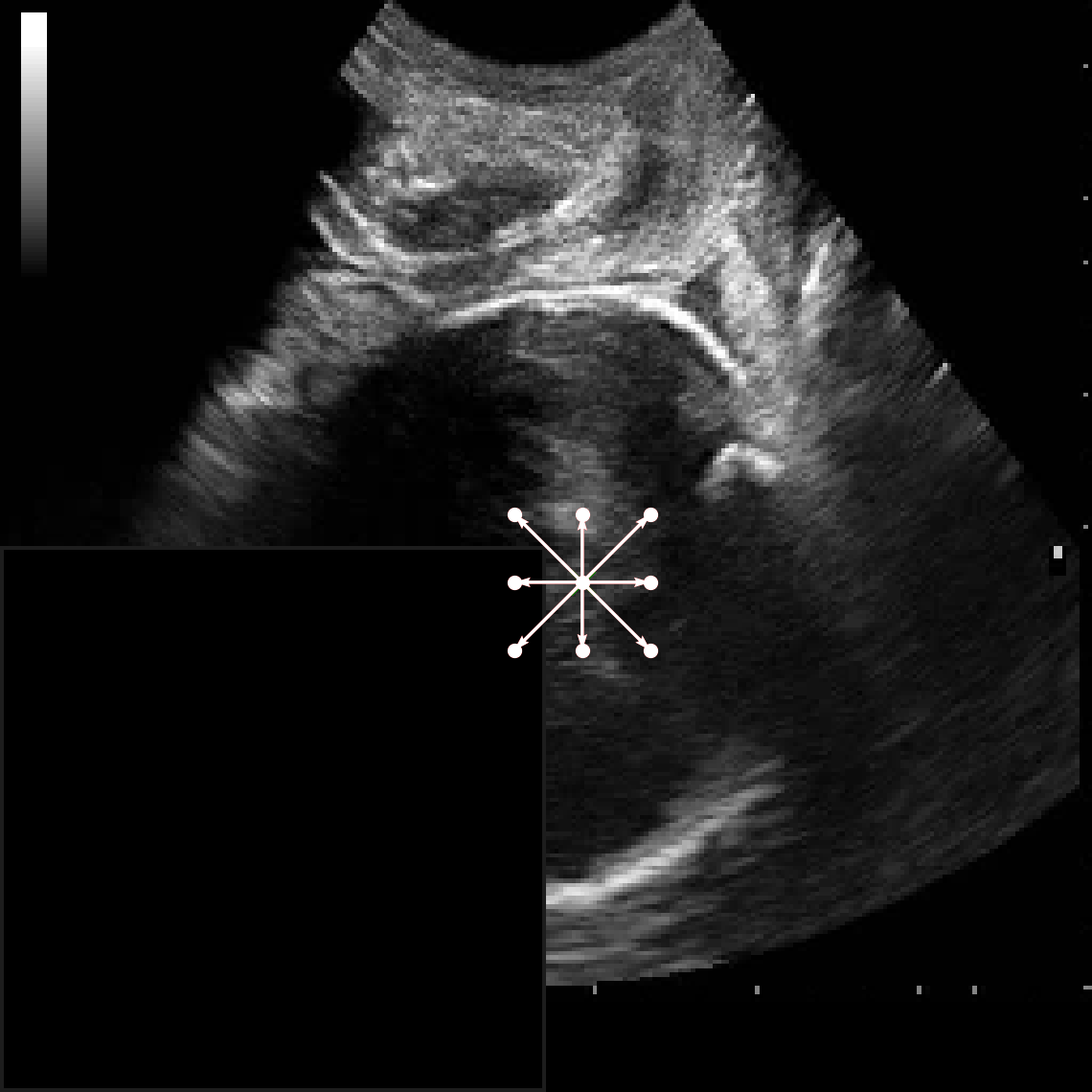} &
  \includegraphics[width=0.18\textwidth]{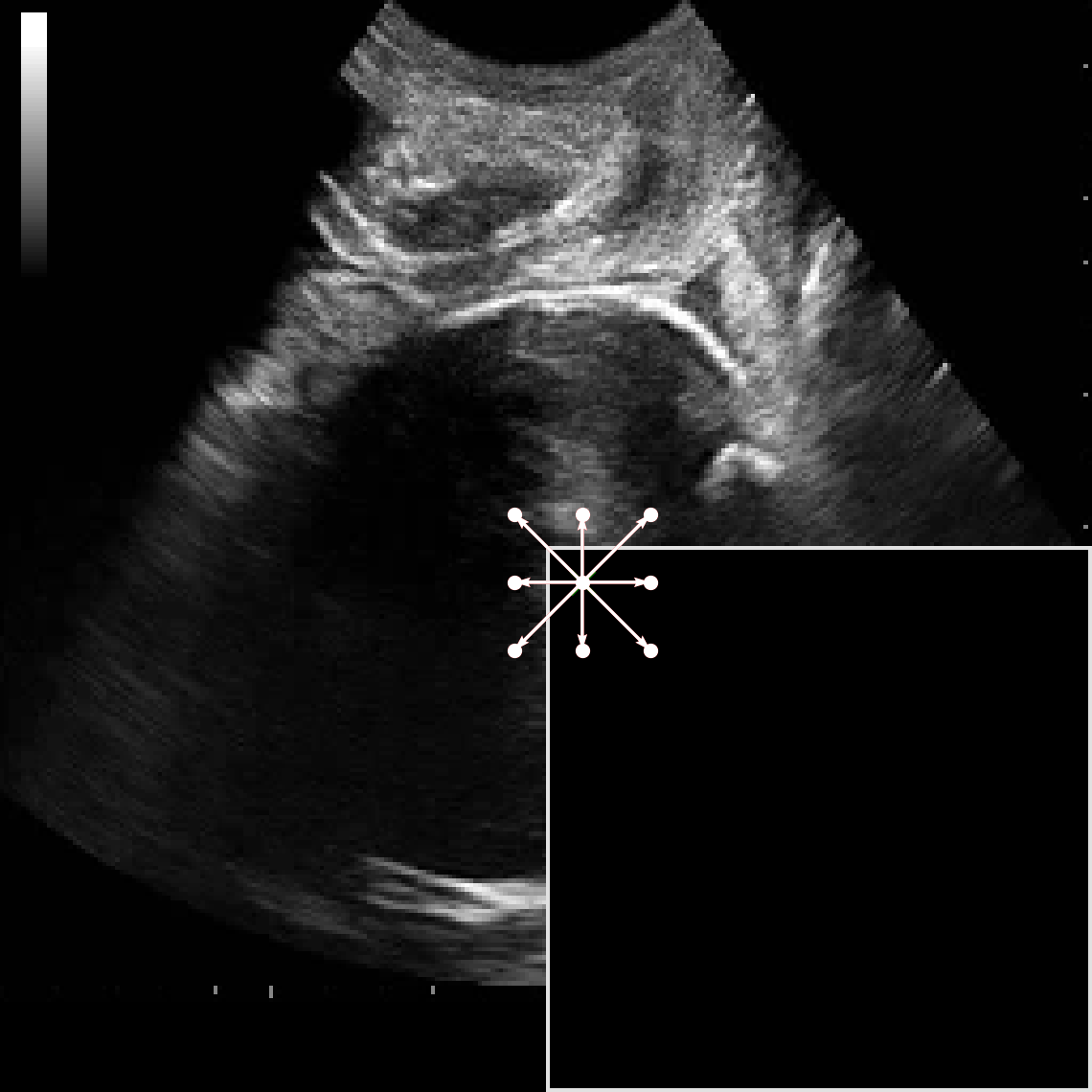} \\
  
  \rotatebox{90}{\hspace{6.5mm} \bf UNet (Masks)} &
  \includegraphics[width=0.18\textwidth]{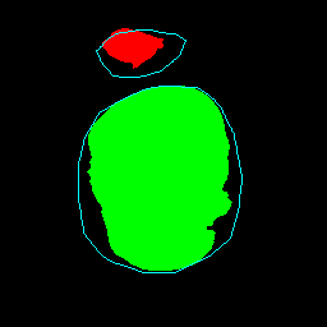} &
  \includegraphics[width=0.18\textwidth]{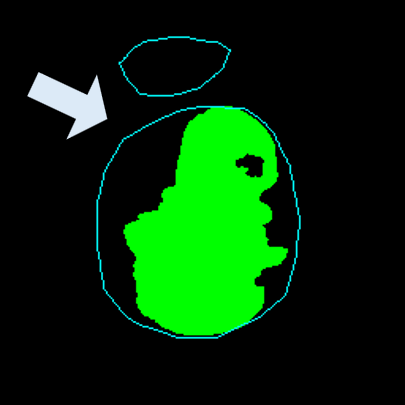} &
  \includegraphics[width=0.18\textwidth]{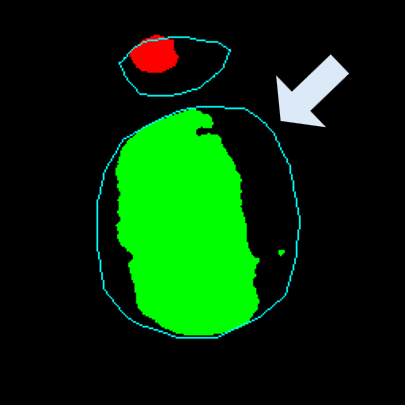} &
  \includegraphics[width=0.18\textwidth]{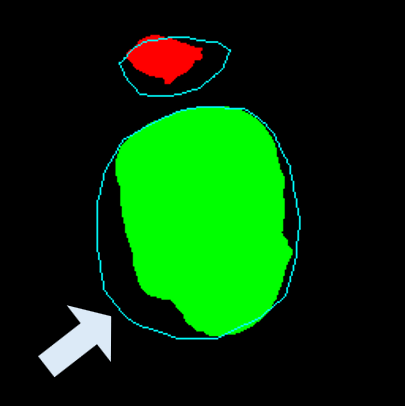} &
  \includegraphics[width=0.18\textwidth]{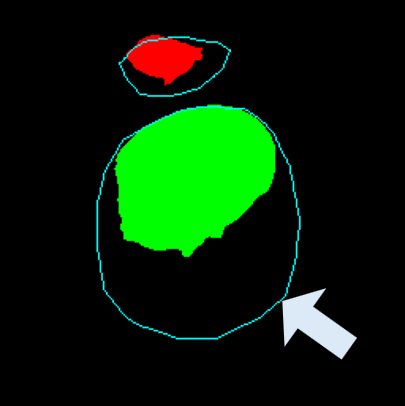} \\
  
  \rotatebox{90}{\hspace{5mm} \bf DAUNet (Offsets)} &
  \includegraphics[width=0.18\textwidth]{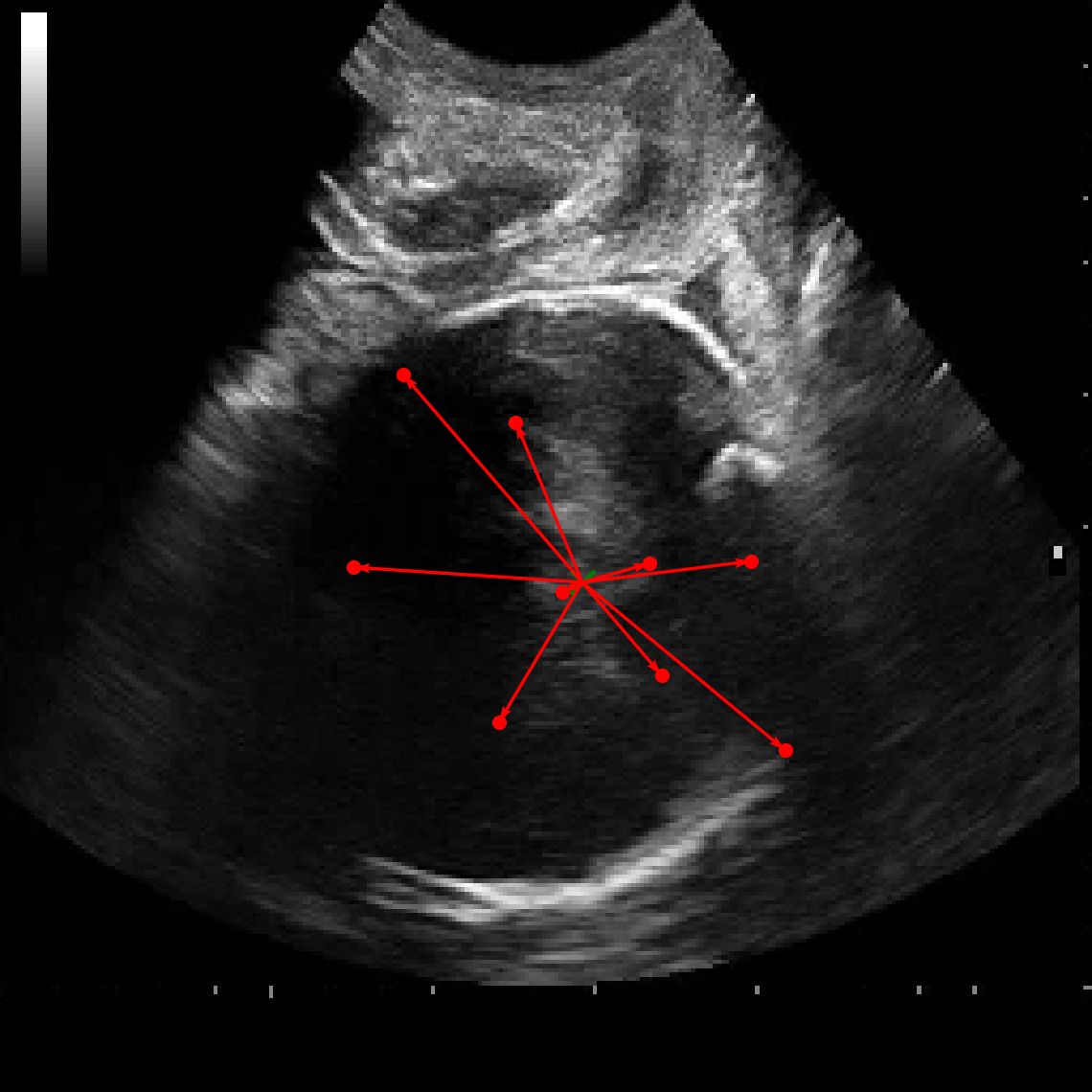} &
  \includegraphics[width=0.18\textwidth]{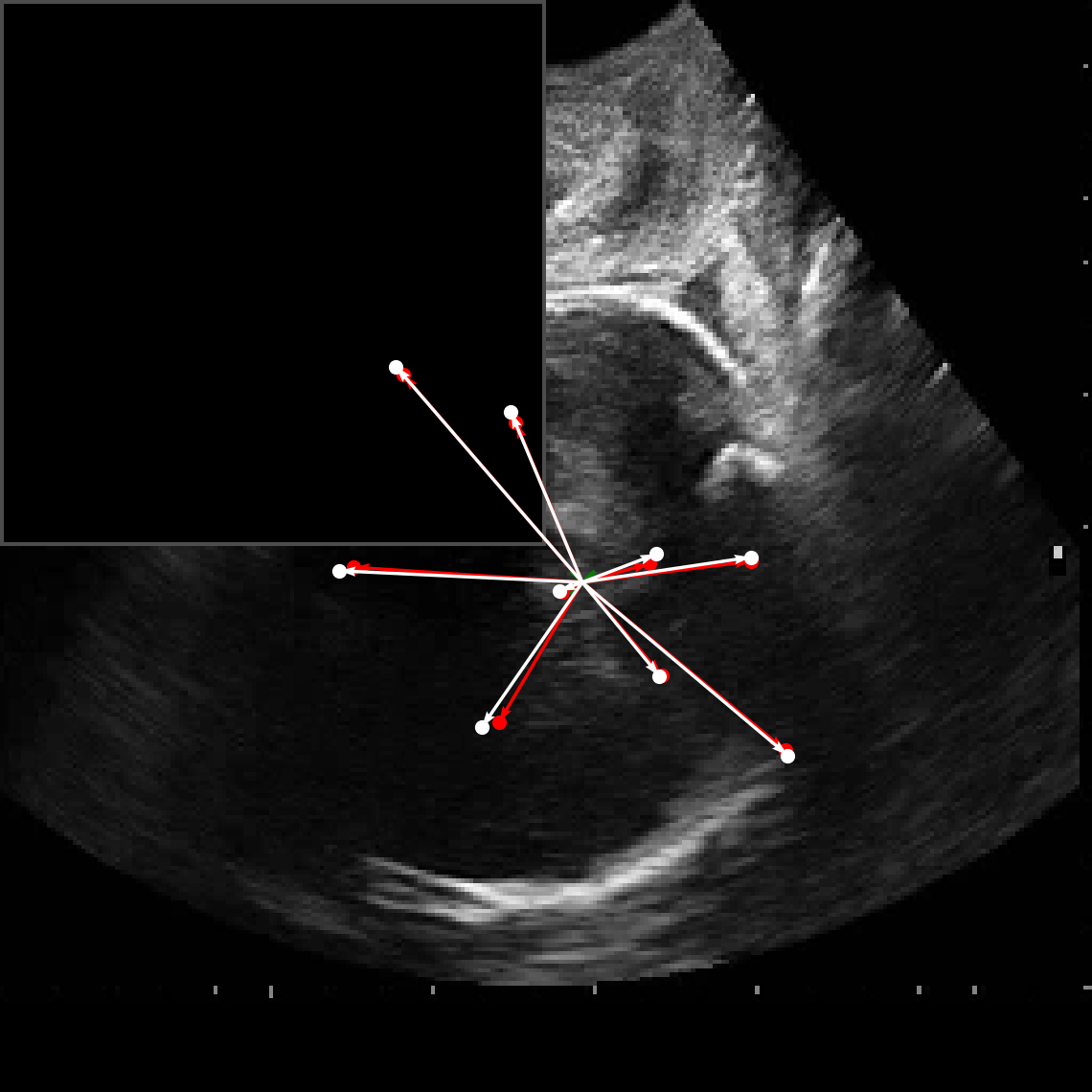} &
  \includegraphics[width=0.18\textwidth]{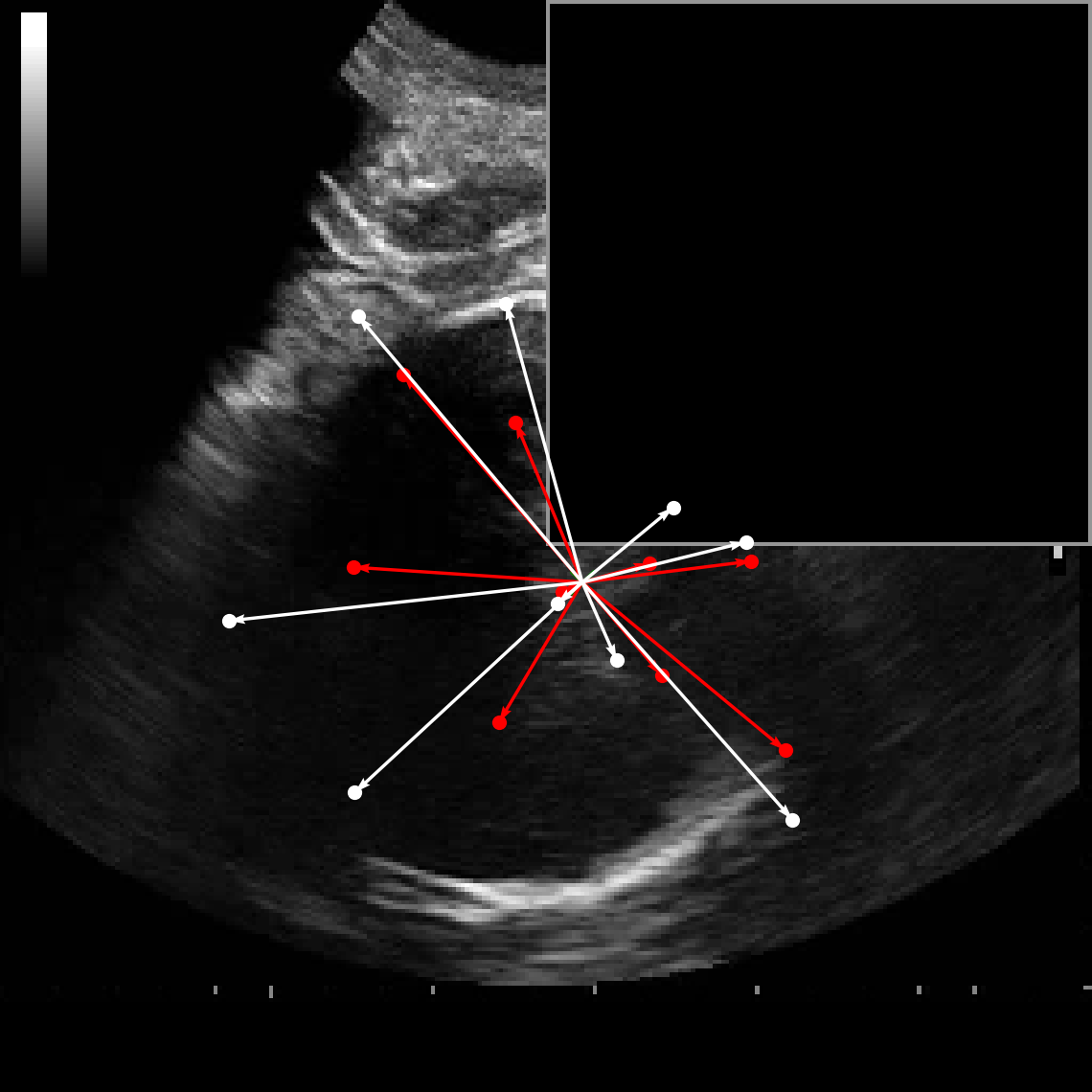} &
  \includegraphics[width=0.18\textwidth]{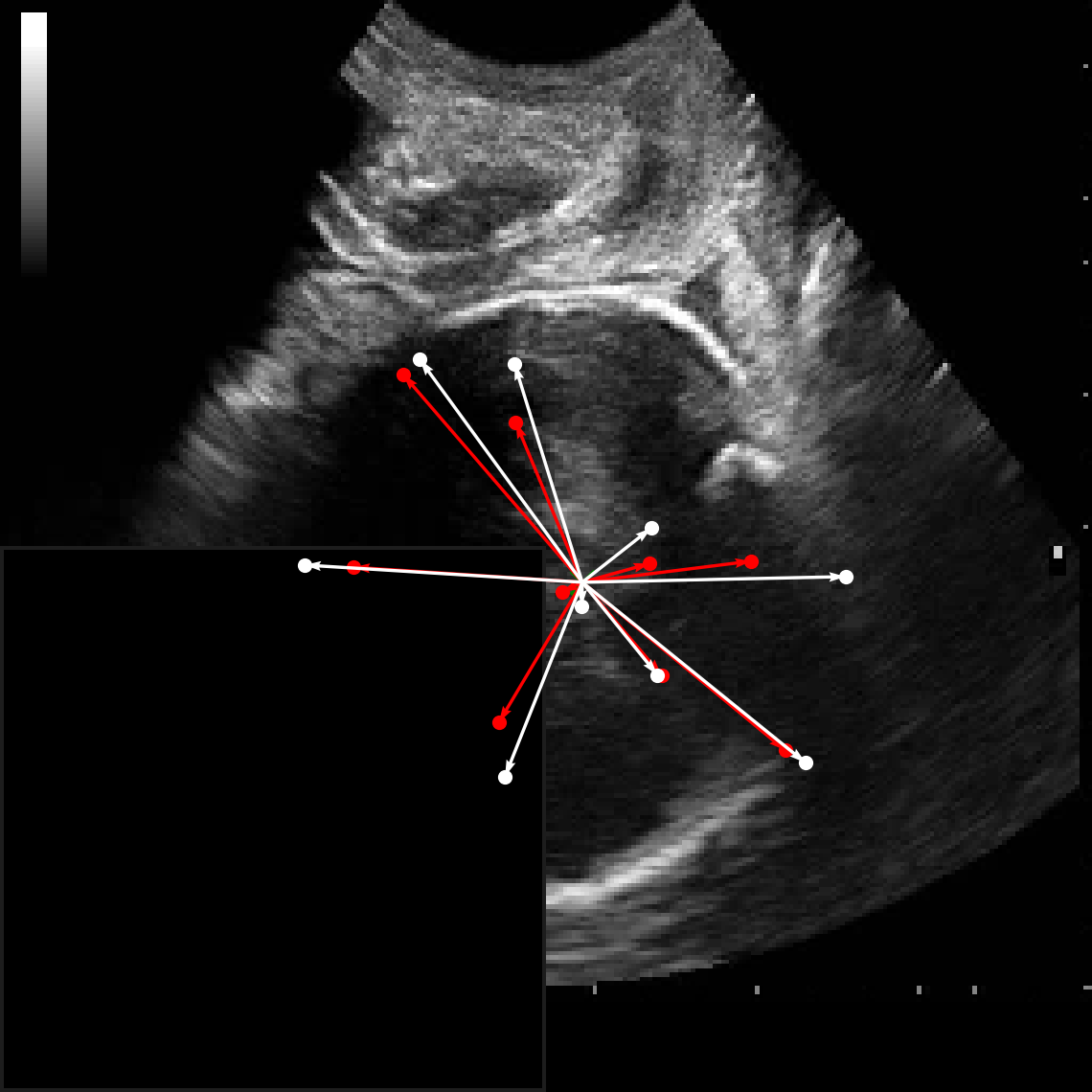} &
  \includegraphics[width=0.18\textwidth]{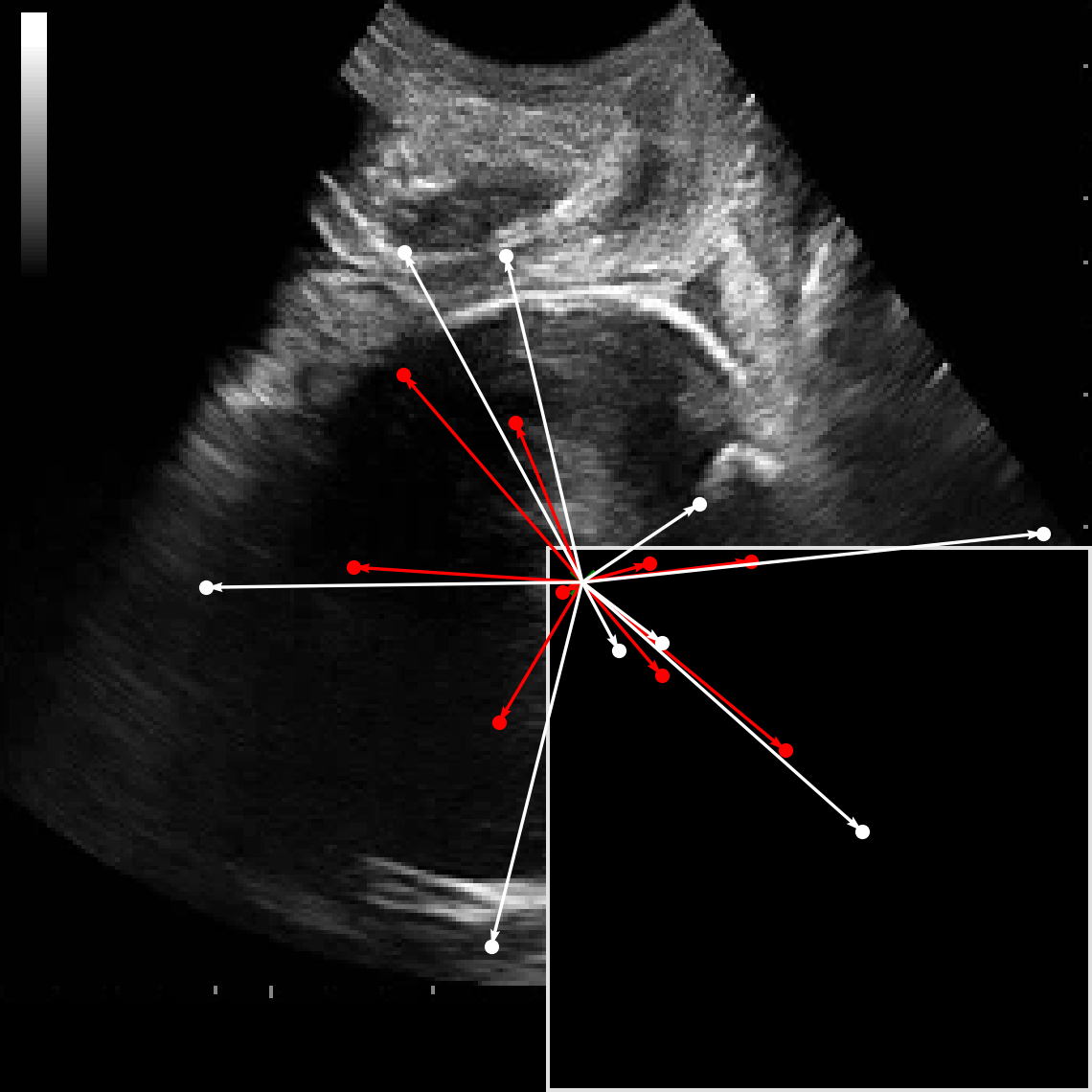} \\
  
  \rotatebox{90}{\hspace{5mm} \bf DAUNet (Masks)} &
  \includegraphics[width=0.18\textwidth]{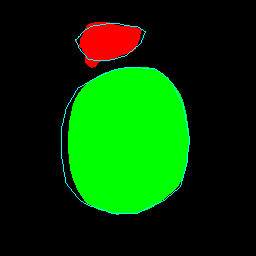} &
  \includegraphics[width=0.18\textwidth]{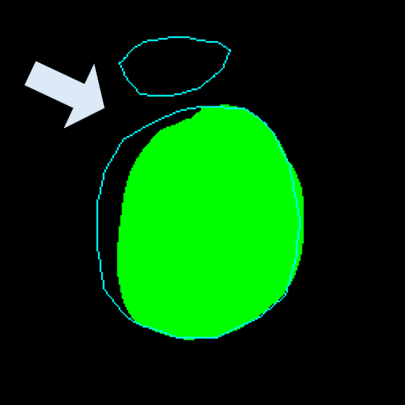} &
  \includegraphics[width=0.18\textwidth]{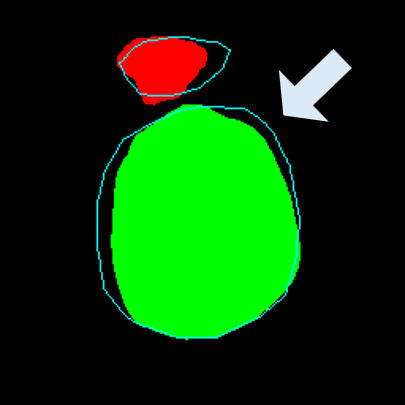} &
  \includegraphics[width=0.18\textwidth]{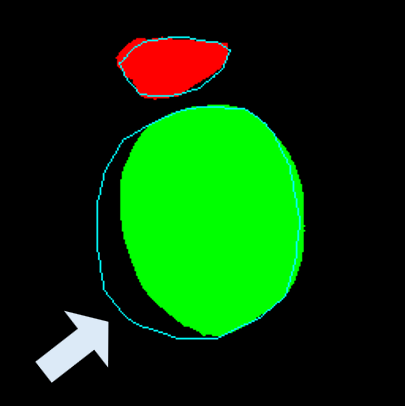} &
  \includegraphics[width=0.18\textwidth]{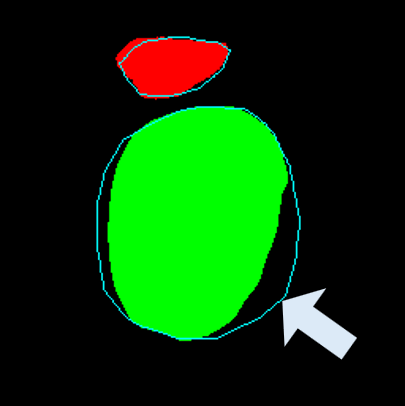} \\
\end{tabular}}
\caption{Robustness to missing context: Comparison of offset maps and segmentation masks generated by UNet and DAUNet for inputs with cropped regions. The first column shows the original image and output; subsequent columns show results for inputs with missing top-left, top-right, bottom-left, and bottom-right patches. DAUNet demonstrates stronger contextual inference and better preservation of structural boundaries despite incomplete input.}
\label{fig:psfh_robustness}
\end{figure*}

In practical clinical settings, medical images are often incomplete or degraded due to occlusions, artifacts, or limited fields-of-view. Therefore, a reliable segmentation model must be robust to missing spatial context and capable of inferring anatomical structures from partially observed inputs. To evaluate this critical property, we conducted a masking experiment using the FH-PS-AoP dataset. In this experiment, we systematically removed one quadrant from the input image, top-left, top-right, bottom-left, or bottom-right, and evaluated the performance of DAUNet compared to the baseline UNet.

Figure~\ref{fig:psfh_robustness} presents a comprehensive visual comparison of the predicted offset maps and segmentation masks under each masking condition. The first column shows the original (unmasked) input along with its corresponding predictions, while subsequent columns depict the results for each of the masked quadrants.

DAUNet demonstrates markedly higher resilience to missing context compared to UNet. Its predicted offset maps remain dense and structured, with vectors that preserve anatomical directionality even when significant regions of the input are absent. This behavior reflects DAUNet’s ability to infer context from the remaining visual cues. In contrast, UNet exhibits sparse or disoriented offsets in the masked scenarios, indicating reduced spatial awareness and compromised localization.

To better understand this phenomenon, we analyzed the receptive fields of both models. For a representative pixel, we visualized its corresponding receptive area contributing to the output. In UNet, the receptive field is fixed and grid-constrained (refer to second row of Figure~\ref{fig:psfh_robustness}), making it sensitive to occlusions. On the other hand, DAUNet leverages deformable convolutions to dynamically adjust its receptive field based on the visible content. This adaptability is evident in the red and white arrows in the fourth column of Figure~\ref{fig:psfh_robustness}, where DAUNet modifies its offset patterns to account for the masked input.

The segmentation masks in third and fifth rows of Figure~\ref{fig:psfh_robustness}, further substantiate these findings. DAUNet consistently produces anatomically plausible segmentations of both the fetal head (green) and pubic symphysis (red), with minimal degradation even under 25\% missing input. The output contours remain smooth, accurate, and well-aligned with ground truth boundaries. In contrast, UNet's performance deteriorates noticeably, with fragmented or distorted segmentations, especially around the fetal head, as indicated by white arrows in the figure.

These results highlight the efficacy of DAUNet's architectural innovations. The proposed combination of deformable convolution and SimAM attention allows the model to effectively reason over the visible context and compensate for spatial omissions. This robustness to incomplete inputs makes DAUNet well-suited for deployment in real-world medical environments, where noise, occlusions, and partial data are common challenges.

\subsection{Ablation Study}

\new{We conduct ablation studies on \emph{both} datasets used in this work to analyze the contribution of each architectural component under different segmentation characteristics. The FH-PS-AoP dataset mainly contains large anatomical structures (fetal head and pubic symphysis) in ultrasound, whereas the FUMPE dataset contains smaller and more challenging embolic targets in CT. Reporting ablations on both datasets helps quantify the effectiveness of the proposed modules across diverse target scales and imaging modalities.}

\subsubsection{Impact of Deformable Convolution-Based Bottleneck}

\begin{table*}[!ht]
\centering
\caption{\new{Combined ablation study on FH-PS-AoP (large anatomical structures) and FUMPE (small lesion targets), analyzing the effect of deformable convolution placement.}}
\label{tab:ablation_study_deformable_placement}

\begin{tabularx}{\linewidth}{>{\centering\arraybackslash}p{0.28\linewidth}|YYY|YY}

\hline \hline
\multirow{2}{*}{\bf Deformable Placement} &
\multicolumn{3}{c|}{\bf FH-PS-AoP (Ultrasound)} &
\multicolumn{2}{c}{\bf FUMPE (CT)} \\ \cline{2-6}
& 
\bf DSC $\uparrow$ & \bf HD95 $\downarrow$ & \bf ASD $\downarrow$ &
\bf DSC $\uparrow$ & \bf ASD $\downarrow$ \\ 
\hline \hline

None &
90.94 & 15.69 & 4.40 & 
85.98 & 4.46 \\ \hline

Encoder &
90.67 & 16.67 & 4.68 & 
77.83 & 4.48 \\ \hline

Decoder &
91.49 & 14.68 & 4.24 & 
72.02 & 4.97 \\ \hline

Proposed (Bottleneck) &
\bf 91.95 & \bf 12.83 & \bf 3.93 &
\bf 88.80 & \bf 4.10 \\ \hline
\hline
\end{tabularx}
\end{table*}

\new{To systematically assess the effect of deformable convolution placement, we evaluate several architectural variants in which deformable convolutions are inserted at different stages of the network, including the encoder, decoder, and bottleneck. The bottleneck represents the lowest-resolution stage of the network, where high-level semantic representations with large receptive fields are formed. As suggested in \cite{ibrahim2025improving}, introducing deformable convolutions at this stage allows the model to adapt its receptive fields to global geometric variations while avoiding excessive computational overhead.}

\new{As shown in Table~\ref{tab:ablation_study_deformable_placement}, placing deformable convolutions in the bottleneck consistently yields the best performance across both datasets. On FH-PS-AoP, the proposed configuration achieves the highest Dice score and the lowest HD95 and ASD, indicating more accurate and better-aligned boundary localization for large anatomical structures. More importantly, the same configuration generalizes effectively to FUMPE, improving segmentation accuracy on small and challenging lesion targets.}

\new{In contrast, inserting deformable convolutions in the encoder or decoder does not provide comparable gains and, in some cases, leads to performance degradation. This behavior suggests that early or late-stage spatial deformation modeling may interfere with low-level feature extraction or fine-grained decoding. Overall, these results justify the design choice of placing deformable convolutions exclusively in the bottleneck to achieve an optimal balance between accuracy, robustness, and computational efficiency.}

\subsubsection{Impact of Parameter-Free Attention (SimAM)}

\begin{table*}[!ht]
\centering
\caption{\new{Combined ablation study on FH-PS-AoP (large anatomical structures) and FUMPE (small lesion targets), analyzing the effect of SimAM placement when using the proposed deformable bottleneck.}}
\label{tab:ablation_study_combined_simam_place}
\begin{tabularx}{\linewidth}{>{\centering\arraybackslash}p{0.28\linewidth}|YYY|YY}
\hline \hline
\multirow{2}{*}{\bf SimAM Placement} &
\multicolumn{3}{c|}{\bf FH-PS-AoP (Ultrasound)} &
\multicolumn{2}{c}{\bf FUMPE (CT)} \\ \cline{2-6}

 & 
\bf DSC $\uparrow$ & \bf HD95 $\downarrow$ & \bf ASD $\downarrow$ &
\bf DSC $\uparrow$ & \bf ASD $\downarrow$ \\ 
\hline \hline

Bottleneck only & 
91.00 & 14.31 & 4.29 &
86.11 & 4.52 \\ \hline

Bottleneck + Encoder + Decoder & 
91.26 & 15.65 & 4.36 &
84.77 & 8.78 \\ \hline

Proposed (Bottleneck + Skip) & 
\bf 91.95 & \bf 12.83 & \bf 3.93 &
\bf 88.80 & \bf 4.10 \\ \hline
\hline
\end{tabularx}
\end{table*}

\new{To investigate the role of parameter-free attention, we evaluate multiple SimAM placement strategies while keeping the deformable bottleneck fixed. SimAM is applied at different stages of the network, including the bottleneck alone, a dense configuration spanning the encoder and decoder, and the proposed configuration that integrates SimAM within the bottleneck and skip connections.}

\new{As shown in Table~\ref{tab:ablation_study_combined_simam_place}, the proposed placement consistently yields the best performance across both datasets. On FH-PS-AoP, integrating SimAM into the bottleneck and skip connections results in the highest Dice score and the lowest HD95 and ASD, indicating improved spatial saliency and boundary refinement for large anatomical structures. Importantly, the same configuration generalizes effectively to FUMPE, delivering the highest Dice score and lowest ASD when segmenting small and challenging lesion targets.}

\new{In contrast, applying SimAM only at the bottleneck provides limited gains, while densely inserting SimAM across both encoder and decoder degrades performance, particularly on FUMPE, suggesting over-suppression of fine-grained features. These results demonstrate that selective attention modulation along skip pathways is critical for preserving spatial detail while enhancing feature discrimination. Overall, the proposed SimAM placement offers an optimal balance between accuracy, robustness, and architectural simplicity, without introducing additional trainable parameters.}

\subsubsection{Combined Contribution Analysis}

\begin{table*}[!ht]
\centering
\caption{\new{Combined ablation study on FH-PS-AoP (ultrasound) and FUMPE (CT) datasets, evaluating the joint contribution of the deformable bottleneck and SimAM attention.}}
\label{tab:ablation_study_combined}

\begin{tabularx}{\linewidth}{>{\centering\arraybackslash}p{0.18\linewidth}|
                                >{\centering\arraybackslash}p{0.10\linewidth}|
                                YYY|YY}

\hline \hline
\multirow{2}{*}{\bf Deformable Bottleneck} & 
\multirow{2}{*}{\bf SimAM} & 
\multicolumn{3}{c|}{\bf FH-PS-AoP (Ultrasound)} &
\multicolumn{2}{c}{\bf FUMPE (CT)} \\ \cline{3-7}

 &  & 
\bf DSC $\uparrow$ & \bf HD95 $\downarrow$ & \bf ASD $\downarrow$ &
\bf DSC $\uparrow$ & \bf ASD $\downarrow$ \\ 
\hline \hline

\ding{55} & \ding{55} & 
89.97 & 17.67 & 5.00 &
77.91 & 6.03 \\ \hline

\ding{55} & \ding{51} & 
90.94 & 15.69 & 4.40 &
85.98 & 5.46 \\ \hline

\ding{51} & \ding{55} & 
91.00 & 14.31 & 4.29 &
86.11 & 4.52 \\ \hline

\ding{51} & \ding{51} & 
\bf 91.95 & \bf 12.83 & \bf 3.93 &
\bf 88.80 & \bf 4.10 \\ \hline \hline
\end{tabularx}
\end{table*}

The combined ablation results in Table~\ref{tab:ablation_study_combined} demonstrate that the deformable bottleneck and SimAM attention contribute complementary benefits to the overall segmentation performance. Individually, each component improves accuracy and boundary localization over the baseline configuration; however, their joint integration yields the most consistent and substantial gains across all evaluation metrics.

On FH-PS-AoP, incorporating both components leads to the highest Dice score and the lowest HD95 and ASD, indicating improved structural delineation and boundary precision for large anatomical regions. \new{More importantly, the same configuration generalizes effectively to FUMPE, where it achieves the best performance on small and challenging lesion targets. This consistency across datasets with markedly different target scales highlights the robustness of the proposed design.}

Overall, these results confirm that the deformable bottleneck enhances geometric adaptability at the feature level, while SimAM refines spatial saliency without increasing model complexity. Their combination yields a synergistic effect that improves segmentation accuracy, robustness, and generalization across modalities, validating the architectural choices adopted in DAUNet.

\subsubsection{Comparison with Alternative Attention Mechanisms}

\begin{table*}[!ht]
\centering
\caption{\new{Combined ablation study comparing different attention mechanisms on FH-PS-AoP (ultrasound) and FUMPE (CT) datasets using the proposed deformable bottleneck.}}
\label{tab:ablation_study_different_attention}

\begin{tabularx}{\linewidth}{>{\centering\arraybackslash}p{0.28\linewidth}|YYY|YY|YY}

\hline \hline
\multirow{2}{*}{\bf Attention} &
\multicolumn{3}{c|}{\bf FH-PS-AoP (Ultrasound)} &
\multicolumn{2}{c|}{\bf FUMPE (CT)} &
\multirow{2}{*}{\bf Param (M) $\downarrow$} &
\multirow{2}{*}{\bf FPS $\uparrow$} \\ \cline{2-6}

 & 
\bf DSC $\uparrow$ & \bf HD95 $\downarrow$ & \bf ASD $\downarrow$ &
\bf DSC $\uparrow$ & \bf HD95 $\downarrow$ &
 &  \\ 
\hline \hline

None &
91.00 & 14.31 & 4.29 &
86.11 & 4.52 &
21.07 & \bf 164.75 \\ \hline

CBAM &
91.67 & 13.65 & 4.04 &
86.32 & 5.21 &
21.29 & 129.03 \\ \hline

ECA &
91.73 & 16.39 & 4.39 &
83.64 & 6.24 &
21.07 & 164.14 \\ \hline

SimAM &
\bf 91.95 & \bf 12.83 & \bf 3.93 &
\bf 88.80 & \bf 4.10 &
\bf 21.07 & 144.33 \\ \hline
\hline
\end{tabularx}
\end{table*}

\new{To further assess the impact of attention design, we compare SimAM with representative lightweight and parameterized attention mechanisms, including Convolutional Block Attention Module (CBAM)~\cite{woo2018cbam} and Efficient Channel Attention (ECA)~\cite{wang2020eca}, while keeping the deformable bottleneck fixed. This analysis isolates the effect of attention formulation on both segmentation performance and computational efficiency.}

\new{As shown in Table~\ref{tab:ablation_study_different_attention}, SimAM consistently achieves the best overall performance across both datasets. On FH-PS-AoP, SimAM yields the highest Dice score and the lowest HD95 and ASD, indicating superior boundary localization. More importantly, SimAM generalizes more effectively to FUMPE, outperforming CBAM in both Dice score and boundary accuracy for small lesion targets.}

\new{Although CBAM improves accuracy over the no-attention baseline, it introduces additional parameters and significantly reduces inference speed, reflecting the overhead of its channel and spatial attention branches. ECA maintains high efficiency but provides limited accuracy gains and does not generalize consistently across datasets. In contrast, SimAM achieves a favorable balance between accuracy and efficiency by refining spatial saliency without introducing extra trainable parameters. These results further justify the choice of SimAM as the attention mechanism in DAUNet.}

\section{Discussion}\label{sec:discussion}

\new{The experiments on FH-PS-AoP (ultrasound) and FUMPE (CT) show that DAUNet provides a favorable balance between segmentation accuracy, boundary precision, and efficiency. On FH-PS-AoP, DAUNet achieves competitive DSC across FH, PS, and PSFH while consistently delivering the lowest HD95 values (Table~\ref{tab:table5_segmentation}), indicating more reliable boundary localization in challenging ultrasound conditions (e.g., speckle noise and weak edges). On FUMPE, DAUNet attains the best overall performance among the compared methods in both overlap and boundary quality (Table~\ref{tab:fumpe_results}), demonstrating that the proposed design generalizes well to small and low-contrast embolic targets. Prior work has shown that leveraging richer contextual cues and auxiliary priors can be particularly beneficial for detecting small and subtle targets in volumetric medical imaging~\cite{ju2025improving}.}

\new{The ablation studies provide insights into why the proposed components are effective. First, placing deformable convolution in the bottleneck yields the strongest and most consistent gains (Table~\ref{tab:ablation_study_deformable_placement}), supporting the design choice of learning spatially adaptive receptive fields at the most semantic stage of the network without disturbing early feature extraction or late decoding. Second, SimAM is most effective when applied selectively along the bottleneck and skip pathways (Table~\ref{tab:ablation_study_combined_simam_place}), improving saliency-aware feature fusion while preserving spatial detail; in contrast, dense insertion across the encoder and decoder can suppress fine structures and degrade generalization. Finally, when compared with alternative attention mechanisms (Table~\ref{tab:ablation_study_different_attention}), SimAM offers the best accuracy--efficiency trade-off, improving boundary metrics without introducing additional learnable parameters or excessive inference overhead. This observation is consistent with recent trends that encourage collaboratively exploiting complementary representations (e.g., dynamic vs.\ static cues) to improve segmentation robustness~\cite{ju2025collaborative}.}

\new{Overall, DAUNet’s improved boundary robustness and lightweight design make it a practical candidate for real-time and resource-constrained clinical deployment, including obstetric ultrasound guidance and pulmonary embolism assessment workflows.}

\subsection{Clinical Impact and Relevance}

In clinical scenarios, accurate segmentation of fetal head and pubic symphysis in transperineal ultrasound is essential for assessing fetal positioning and guiding delivery decisions~\cite{salomon2019practice}. Errors in segmentation can lead to misjudgment in head station evaluation, potentially increasing the risk of obstructed labor~\cite{memon2013vaginal}. DAUNet's robust performance on the FH-PS-AoP dataset highlights its suitability for integration into real-time obstetric ultrasound systems, potentially assisting clinicians in low-resource or emergency settings.

Similarly, the FUMPE dataset targets the detection of pulmonary embolism (PE), a life-threatening condition that demands prompt diagnosis~\cite{marshall2011diagnosis}. DAUNet demonstrated excellent boundary precision and robustness in segmenting emboli within peripheral pulmonary arteries, which are often missed by conventional methods. By providing accurate and fast segmentation, DAUNet could be integrated into clinical decision support systems (CDSS) or computer-aided diagnosis (CAD) tools to assist radiologists in detecting PE more efficiently.

Moreover, the architectural design of DAUNet, featuring deformable convolutions and parameter-free attention, ensures that the model generalizes well across different imaging modalities and anatomical structures. This adaptability is clinically valuable, as it reduces the need for extensive retraining and fine-tuning for other challenging applications, such as tumor boundary delineation in oncology or vessel segmentation in cardiology.

From a translational perspective, DAUNet's low parameter count and fast inference speed make it an excellent candidate for deployment on portable ultrasound machines and embedded systems, thereby enabling point-of-care diagnostics and remote healthcare delivery. This aligns with current trends in mobile health (mHealth), where AI models are increasingly used to bring expert-level diagnostics to underserved populations~\cite{davenport2019potential}.

\section{Conclusion}\label{sec:conclusion}
In this paper, we introduced DAUNet, a lightweight and efficient UNet-based segmentation architecture that integrates Deformable V2 Convolutions and SimAM, a parameter-free attention mechanism, to enhance spatial adaptability and feature selectivity without increasing model complexity. Through extensive evaluation on two challenging datasets, including FH-PS-AoP for ultrasound-based fetal head and pubic symphysis segmentation, and FUMPE for pulmonary embolism detection from CT, DAUNet demonstrated superior performance in terms of segmentation accuracy, boundary preservation, and parameter efficiency compared to existing state-of-the-art models. Ablation studies confirmed the individual and combined contributions of the proposed components, while robustness analysis under missing context scenarios highlighted the model's ability to infer anatomical structure even with incomplete input. These results underscore DAUNet’s potential for deployment in real-world clinical environments, particularly where computational resources are limited or image quality is compromised. Future work will explore extending the framework to multi-modal and 3D imaging, improving cross-domain generalization through domain adaptation, and further optimizing the model for real-time inference on edge devices.

\section*{Acknowledgment}
Shujaat Khan acknowledges the support from the King Fahd University of Petroleum \& Minerals (KFUPM) under Early Career Research Grant no. EC241027. Additionally, the support from KFUPM Ibn Battuta Global Scholarship Program Grant No: ISP241-COE-872 is gratefully acknowledged.


\begin{IEEEbiography}[{\includegraphics[width=1in,height=1.25in,clip,keepaspectratio]{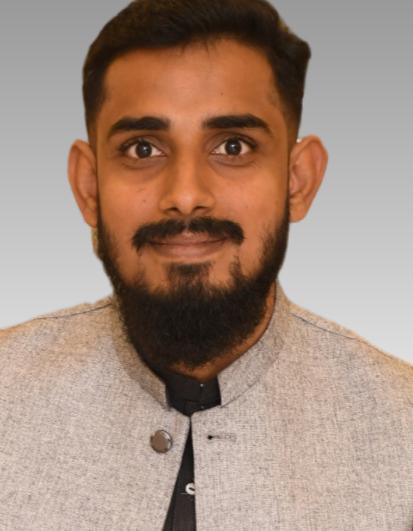}}]{Adnan Munir} 
received the B.Sc. degree in Computer Engineering from COMSATS University Islamabad, Pakistan, in 2019, and the M.Sc. degree in Computer Engineering from King Fahd University of Petroleum and Minerals (KFUPM), Dhahran, Saudi Arabia, in 2024. From 2019 to 2021, he worked as a Research and Development Engineer at E-wall, Pakistan. Since 2021, he has been affiliated with KFUPM as a Graduate Teaching Assistant in the Department of Computer Engineering.

His research interests include medical imaging, deep learning, computer vision, adversarial attacks, out-of-distribution detection, embedded systems, and wireless ad hoc networks. He has authored or co-authored several publications in peer-reviewed journals and international conferences in these areas.
\end{IEEEbiography}

\begin{IEEEbiography}[{\includegraphics[width=1in,height=1.25in,clip,keepaspectratio]{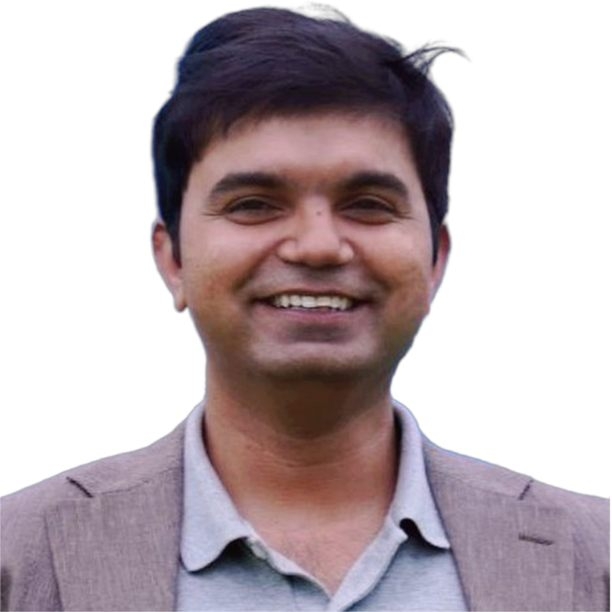}}]{Muhammad Shahid Jabbar} earned his Ph.D. in Computer Engineering from Sungkyunkwan University, Suwon, South Korea, in 2022. He was a Researcher at the H-Lab and the Media System Lab at SKKU. His doctoral research focused on developing modern deep learning solutions using multimodal data for convergence applications involving multi-sensory interaction and cross-modular associations, with particular emphasis on accessibility for visually impaired users. He is currently a Post-Doctoral Fellow at the SDAIA–KFUPM Joint Research Center for Artificial Intelligence, King Fahd University of Petroleum \& Minerals (KFUPM), Dhahran, KSA. Prior to joining KFUPM, he worked as Senior AI Researcher at Radisen Co. Ltd., Seoul, South Korea, where he contributed to AI-powered digital radiography solutions for medical image analysis. His research interests include computer vision, diffusion models, multimodal learning, and inverse problems, with a focus on biomedical and human-centered artificial intelligence.
\end{IEEEbiography}

\begin{IEEEbiography}[{\includegraphics[width=1in,height=1.25in,clip,keepaspectratio]{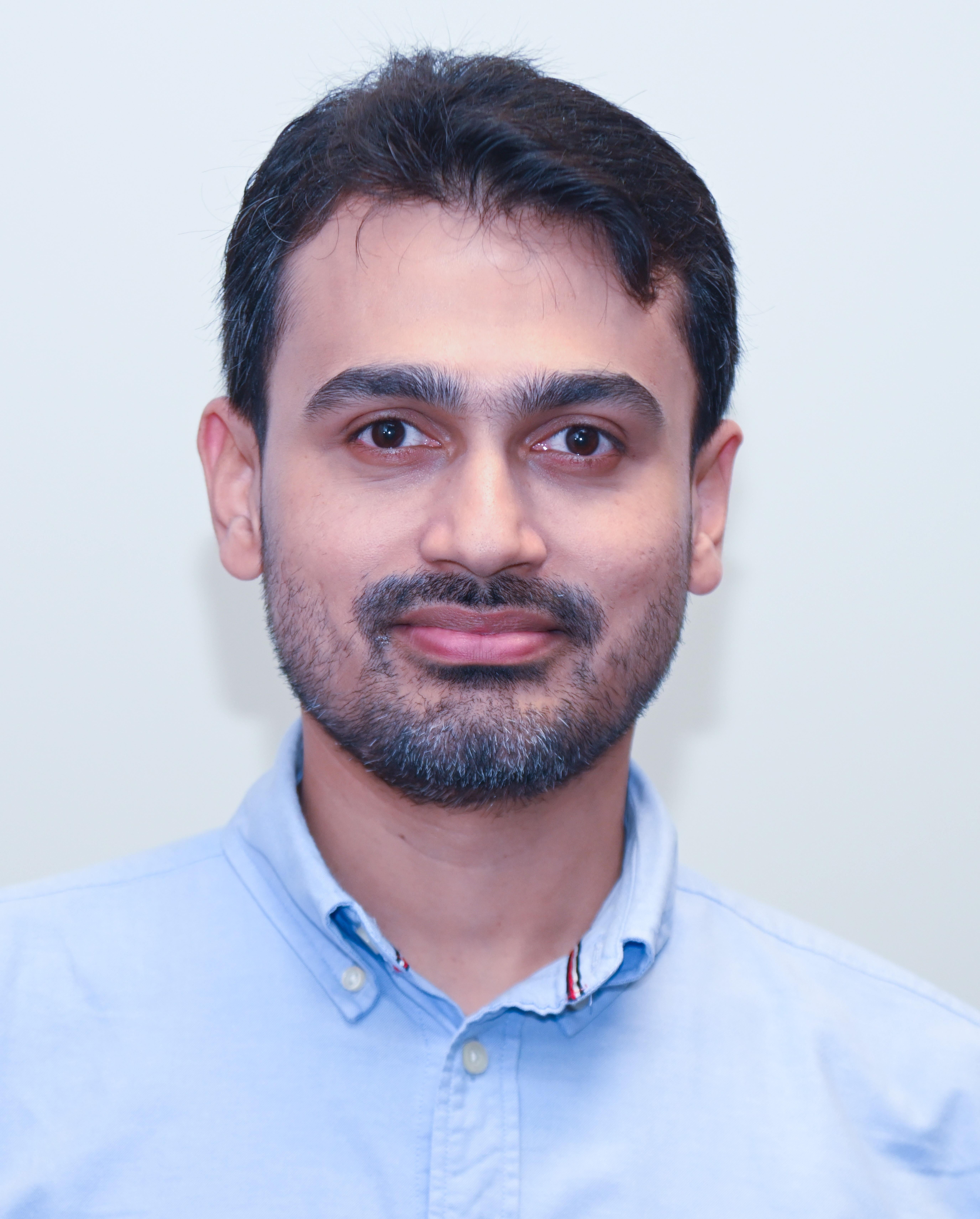}}]{Shujaat Khan} is an Assistant Professor in the Department of Computer Engineering and a Fellow at the Saudi Data and AI Authority (SDAIA) and King Fahd University of Petroleum \& Minerals (KFUPM) under the SDAIA-KFUPM Joint Research Center for Artificial Intelligence (JRC-AI) at KFUPM, Dhahran, KSA. Prior to joining KFUPM, he was a Senior AI Scientist at Digital Technology \& Innovation, Siemens Medical Solutions USA, Inc. He earned his Ph.D. from the Department of Bio and Brain Engineering at the Korea Advanced Institute of Science and Technology (KAIST), Daejeon, South Korea, in 2022. He was a researcher with Synergistic Bioinformatics (SynBi) and the Bio Imaging, Signal Processing Learning (BISPL) Labs at KAIST. His research interests include machine learning, optimization, inverse problems, and signal processing, with a focus on biomedical and bioinformatics applications. 
\end{IEEEbiography}

\end{document}